\documentclass[10pt,twocolumn,letterpaper]{article}

\usepackage{iccv}
\usepackage{times}
\usepackage{cite}
\usepackage{epsfig}
\usepackage{graphicx}

\usepackage{amsmath}
\usepackage{stfloats}
\usepackage[colorlinks,linkcolor=red]{hyperref}
\usepackage{amssymb}
\usepackage{algpseudocode}
\usepackage{amsmath}
\usepackage{picins}
\usepackage{stfloats}
\usepackage{algorithm}
\usepackage{algorithmicx}
\usepackage{algpseudocode}
\usepackage{amsmath}

\iccvfinalcopy 

\begin{document}

\title{Coherent Semantic Attention for Image Inpainting}
\author{Hongyu Liu\\
\and
Bin Jiang\\
\and
Yi Xiao\\
\and
Chao Yang\\
\and
College of Computer Science and Electronic Engineering\\
Hunan University\\
}

\maketitle

\begin{figure*}[!htbp]
\centering
\includegraphics[scale=0.38]{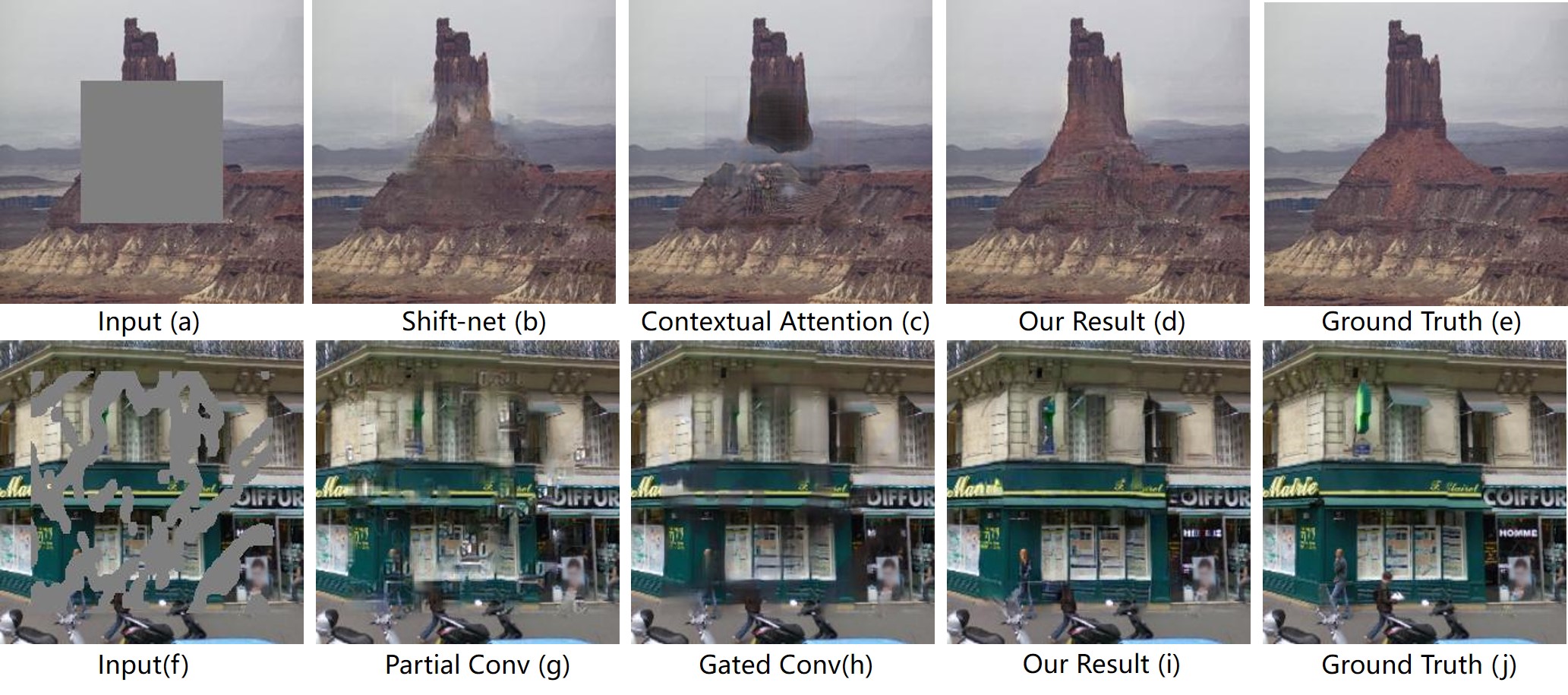}
\caption{Our results compared with Contextual Attention~\cite{8}, Shift-net~\cite{9}, Partial Conv~\cite{38}, and Gated Conv~\cite{39}. First line, from left to right are: image with centering mask, Shift-net~\cite{9}, Contextual Attention~\cite{8}, our model, Ground Truth, respectively. Second line, from left to right are: image with irregular mask, Partial Conv~\cite{38}, Gated Conv~\cite{39}, our model, Ground Truth, respectively. The size of images are 256$\times$256.
   }
\label{img1}
\end{figure*}


\begin{abstract}
The latest deep learning-based approaches have shown promising results for the challenging task of inpainting missing regions of an image. However, the existing methods often generate contents with blurry textures and distorted structures due to the discontinuity of the local pixels. From a semantic-level perspective, the local pixel discontinuity is mainly because these methods ignore the semantic relevance and feature continuity of hole regions. To handle this problem, we investigate the human behavior in repairing pictures and propose a fined deep generative model-based approach with a novel coherent semantic attention (CSA) layer, which can not only preserve contextual structure but also make more effective predictions of missing parts by modeling the semantic relevance between the holes features. The task is divided into rough, refinement as two steps and model each step with a neural network under the U-Net architecture, where the CSA layer is embedded into the encoder of refinement step. To stabilize the network training process and promote the CSA layer to learn more effective parameters, we propose a consistency loss to enforce the both the CSA layer and the corresponding layer of the CSA in decoder to be close to the VGG feature layer of a ground truth image simultaneously. The experiments on CelebA, Places2, and Paris StreetView datasets have validated the effectiveness of our proposed methods in image inpainting tasks and can obtain images with a higher quality as compared with the existing state-of-the-art approaches. The codes and pre-trained models will be available at \url{https://github.com/KumapowerLIU/CSA-inpainting}.
\end{abstract}

\section{Introduction}

Image inpainting is the task to synthesize the missing or damaged parts of a plausible hypothesis, and can be utilized in many applications such as removing unwanted objects, completing occluded regions, restoring damaged or corrupted parts. The core challenge of image inpainting is to maintain global semantic structure and generate realistic texture details for the missing regions.

Traditional works~\cite{1,2,3,18,43} mostly develop texture synthesis techniques to address the problem of hole filling. In~\cite{3}, Barnes et al. propose the Patch-Match algorithm which iteratively searches for the best fitting patches from hole boundaries to synthesize the contents of the missing parts. Wilczkowiak et al.~\cite{43} take further steps and detect desirable search regions to find better match patches. However, these methods fall short of understanding high-level semantics and struggle at reconstructing patterns that are locally unique. In contrast, early deep convolution neural networks based approaches~\cite{4,5,6,7} learn data distribution to capture the semantic information of the image, and can achieve plausible inpainting results. However, these methods fail to effectively utilize contextual information to generate the contents of holes, often leading to the results containing noise patterns.

Some recent studies effectively utilize the contextual information and obtain better inpainting results. These methods can be divided into two types. The first type~\cite{8,9,10} utilizes spatial attention which takes surrounding image features as references to restore missing regions. These methods can ensure the semantic consistency of generated content with contextual information. However, they just focus on rectangular shaped holes, and the results always tend to show pixel discontinuous and have semantic chasm (See in Fig \ref{img1}(b, c)). The second type~\cite{38,39} is to make the prediction of the missing pixels condition on the valid pixels in the original image. These methods can handle irregular holes properly, but the generated contents still meet problems of semantic fault and boundary artifacts (See in Fig \ref{img1}(g, h)). The reason that the above mentioned methods do not work well is because they ignore the semantic relevance and feature continuity of generated contents, which is crucial for the local pixel continuity.

In order to achieve better image restoration effect, we investigate the human behavior in inpainting pictures and find that such process involves two steps as conception and painting to guarantee both global structure consistency and local pixel continuity of a picture. To put it more concrete, a man first observes the overall structure of the image and conceives the contents of missing parts during conception process, so that the global structure consistency of the image can be maintained. Then the idea of the contents will be stuffed into the actual image during painting process. In the painting process, one always continues to draw new lines and coloring from the end nodes of the lines drawn previously, which actually ensures the local pixel continuity of the final result.

Inspired by this process, we propose a coherent semantic attention layer (CSA), which fills in the unknown regions of the image feature maps with the similar process. Initially, each unknown feature patch in the unknown region is initialized with the most similar feature patch in the known regions. Thereafter, they are iteratively optimized by considering the spatial consistency with adjacent patches. Consequently, the global semantic consistency is guaranteed by the first step, and the local feature coherency is maintained by the optimizing step.

Similar to~\cite{8}, we divide the image inpainting into two steps. The first step can be constructed by training a rough network to rough out the missing contents. A refinement network with the CSA layer in encoder guides the second step to refine the rough predictions. In order to make network training process more stable and motivate the CSA layer to learn more effective features, we propose a consistency loss to measure not only the distance between the VGG feature layer and the CSA layer but also the distance between the VGG feature layer and the the corresponding layer of the CSA in decoder. Meanwhile, in addition to a patch discriminator~\cite{11}, we improve the details by introducing a feature patch which is simpler in formulation, faster and more stable for training than conventional one~\cite{16}. Except for the consistency loss, reconstruction loss, and relativistic average LS adversarial loss~\cite{12} are incorporated as constraints to instruct our model to learn meaningful parameters.

We conduct experiments on standard datasets CelebA~\cite{13}, Places2~\cite{14}, and Paris StreetView~\cite{40}. Both the qualitative and quantitative tests demonstrate that our method can generate higher-quality inpainting results than existing ones. (See in Fig \ref{img1}(d, i)).

Our contributions are summarized as follows:
\begin{itemize}
\item We propose a novel coherent semantic attention layer to construct the correlation between the deep features of hole regions. No matter whether the unknown region is irregular or centering, our algorithm can achieve state-of-the-art inpainting results.
\item To enhance the performance of the CSA layer and training stability, we introduce the consistency loss to guide the CSA layer and the corresponding decoder layer to learn the VGG features of ground truth. Meanwhile, a feature patch discriminator is designed and jointed to achieve better predictions.
\item Our approach achieves higher-quality results in comparison with~\cite{8,9,38,39} and generates more coherent textures. Even the inpainting task is completed in two stages, our full network can be trained in an end to end manner.
\end{itemize}

\section{Related Works}
\subsection{Image inpainting}
\begin{figure*}[t]
\centering
\includegraphics[scale=0.51]{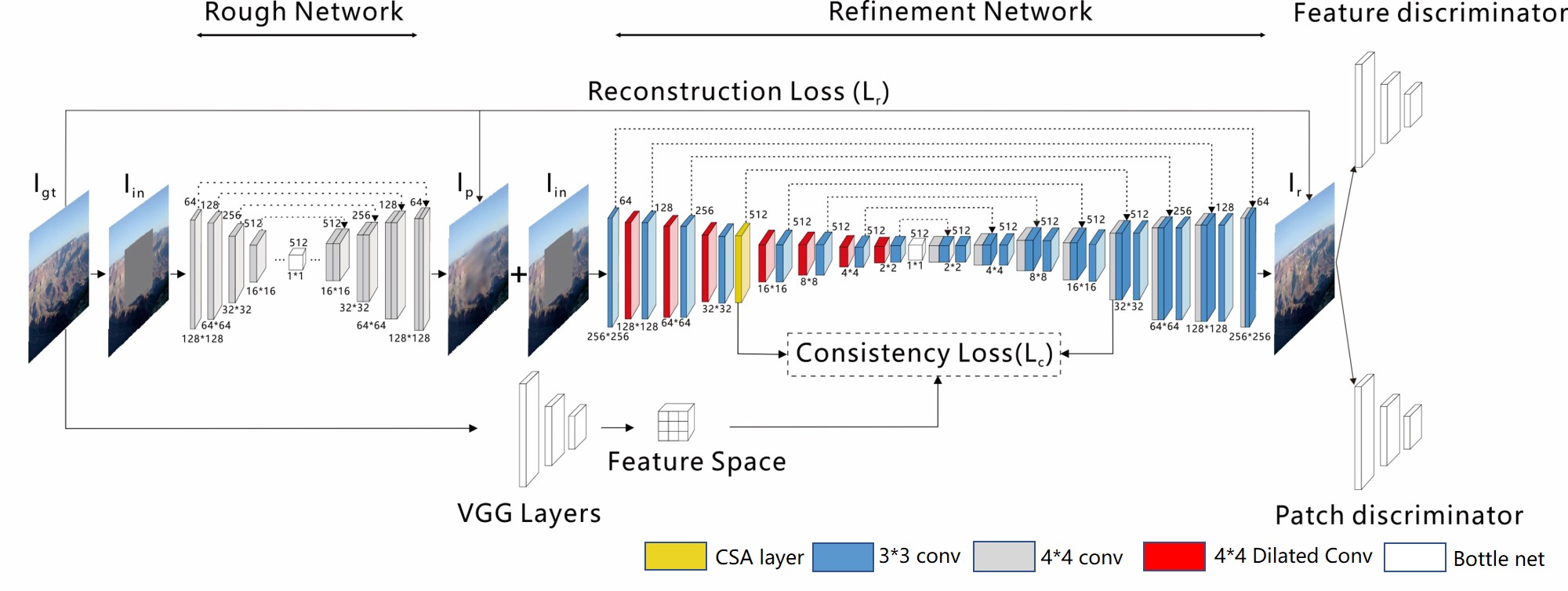}
\caption{The architecture of our model. We add the CSA layer at the resolution of 32$\times32$ in refinement network.}
\label{img2}
\end{figure*}
In the literature, previous image inpainting researches can generally be divided into two categories: Non-learning inpainting approaches and Learning inpainting approaches. The former is traditional diffusion-based or patch-based methods with low-level features. The latter learns the semantics of image to fulfill the inpainting task and generally trains deep convolutional neural networks to infer the content of the missing regions.

Non-learning approaches such as~\cite{1,2,17,18,19,20,21,22,23,24,25,26,27,28} fill in missing regions by propagating neighboring information or copying information from similar patch of the background. Huang et al.~\cite{29} blend the known regions into the target regions to minimize discontinuities. However, searching the best matching known regions is a very expensive operation. To address this challenge, Barnes et al.~\cite{3} propose a fast nearest neighbor field algorithm which promotes the development of image inpainting applications. Though the non-learning approaches work well for surface textures synthesis, they can not generate semantically meaningful content, and are not suitable to deal with large missing regions.

Learning approaches ~\cite{31,32,33,34,35,36,37} often use deep learning and GAN strategy to generate pixels of the hole. Context encoders~\cite{7} firstly train deep neural networks for image inpainting task, which takes the adversarial training ~\cite{41} into a novel encoder-decoder pipeline and outputs prediction of missing regions. However, it performs poorly in generating fine-detailed textures. Soon after that, Iizuka et al.~\cite{4} extend this work and propose local and global discriminators to improve the inpainting quality. However, it requires post processing steps to enforce the color coherency near the hole boundaries. Yang et al.~\cite{30} take the result from context encoders~\cite{7} as input and gradually increase the texture details to get high-resolution prediction. But this approach significantly increases computational costs due to its optimization process. Liu et al.~\cite{38} update the mask in each layer and re-normalize the convolution weights with the mask value, which ensures that the convolution filters concentrate on the valid information from known regions to handle irregular holes. Yu et al.~\cite{39} further propose to learn the mask automatically with gated convolutions, and combine with SN-PatchGAN discriminator to achieve better predictions. However, these methods do not explicitly consider the correlation between valid features, thus resulting in color inconsistency on completed image.
\subsection{Attention based image inpainting}
Recently, the spatial attention based on the relationship between contextual and hole regions is often used for image inpainting tasks. Contextual Attention~\cite{8} proposes a contextual attention layer which searches for a collection of background patches with the highest similarity to the coarse prediction. Yan et al.~\cite{9} introduce a shift-net powered by a shift operation and a guidance loss. The shift operation speculate the relationship between the contextual regions in the encoder layer and the associated hole region in the decoder layer. Song et al.~\cite{10} introduce a patch-swap layer, which replaces each patch inside the missing regions of a feature map with the most similar patch on the contextual regions, and the feature map is extracted by VGG network. Although~\cite{8} has the spatial propagation layer to encourage spatial coherency by the fusion of attention scores, it fails to model the correlations between patches inside the hole regions, which is also the drawbacks of the other two methods. To this end, we proposed our approach to solve this problem and achieve better results, which is detailed in Section 3.

\section{Approach}
Our model consists of two steps: rough inpainting and refinement inpainting. This architecture helps to stabilize training and enlarge the receptive fields as mentioned in~\cite{8}. The overall framework of our inpainting system is shown in Fig \ref{img2}. Let $I_{gt}$ be the ground truth images, $ I_{in}$ be the input to the rough network, the $M$ and $\overline{M}$ denote the missing area and the known area in feature maps respectively. We first get the rough prediction $I_p$ during the rough inpainting process. Then, the refinement network with CSA layer takes the $I_p$ and $I_{in}$ as input pairs to output final result $I_r$. Finally, the patch and feature patch discriminators work together to obtain higher resolution of $I_r$.
\subsection{Rough inpainting}
The input of rough network $I_{in}$ is a 3$\times$256$\times$256 image with center or irregular holes, which is sent to the rough net to output the rough prediction $I_p$. The structure of our rough network is the same as the generative network in~\cite{11}, which is composed of 4$\times$4 convolutions with skip connections to concatenate the features from each layer of encoder and the corresponding layer of decoder. The rough network is trained with the $L_1$ reconstruction loss explicitly.

\subsection{Refinement inpainting}
\subsubsection{refinement network}
We use $I_{p}$ conditioned on $I_{in}$ as input of refinement network that predicts the final result $I_r$. This type of input stacks information of the known areas to urge the network to capture the valid features faster, which is critical for rebuilding the content of hole regions. The refinement network consists of an encoder and a decoder, where skip connection is also adopted similar to rough network. In the encoder, each of the layers is composed of a 3$\times$3 convolution and a 4$\times$4 dilated convolution. The 3$\times$3 convolutions keep the same spatial size while doubling the number of channels. Layers of this size can improve the ability of obtaining deep semantic information. The 4$\times$4 dilated convolutions reduce the spatial size by half and keep the same channel number. The dilated convolutions can enlarge the receptive fields, which can prevent excessive information loss. The CSA layer is embedded in the fourth layer of the encoder. The structure of decoder is symmetrical to the encoder without CSA layer and all 4$\times$4 convolutions are deconvolutions.
\subsubsection{Coherent Semantic Attention}
We believe that it is not enough to only consider the relationship between $M$ and $\overline{M}$ in feature map to reconsturct $M$ similar to~\cite{8,9,10}, because the correlation between generated patches is ignored, which may result in lack of ductility and continuity in the final result.

\begin{figure}[h]\label{eq3}
\begin{center}
\includegraphics[scale=0.27]{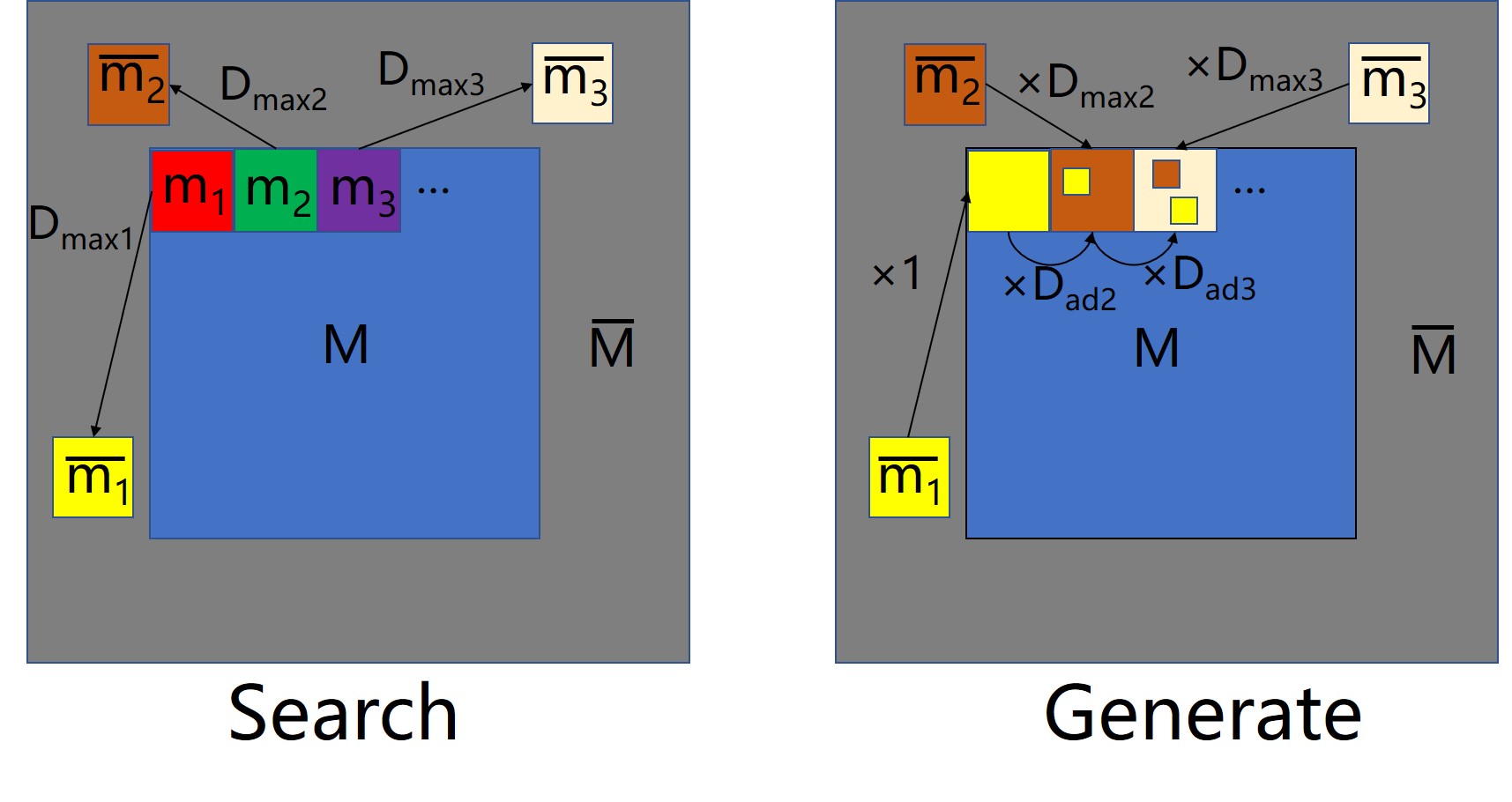}
\end{center}
   \caption{Illustration of the CSA layer. Firstly, each neural patch in the hole $M$ searches for the most similar neural patch on the boundary $\overline{M}$. Then, the previous generated patch and the most similar contextual patch are combined to generate the current one.
}
\label{img3}
\end{figure}

To overcome this limitation, we consider the correlation between generated patches and propose the CSA layer. We take the centering hole as an example: the CSA layer is implemented in two phases: Search and Generate. For each (1$\times$1) generated patch $m_i$ in $M$ ($i\in (1\sim n)$, $n$ is the number of patches), the CSA layer searches the closest-matching neural patch $\overline{m_i}$ in known region $M$ to initialize $m_i$ during the search process. Then we set the $\overline{m_i}$ as a main reference and the previous generated patch $m_{i-1}$ as a secondary information to restore $m_i$ during the generative process. To measure the relevant degree between these patches, the following cross-correlation metric is adopted:
\begin{equation}\label{eq1}
\begin{aligned}
Dmax_i=\frac{<m_i,\overline{m_i}>}{||m_i||.||\overline{m_i}||}
\end{aligned}
\end{equation}
\begin{equation}\label{eq2}
\begin{aligned}
Dad_i =\frac{<m_i,m_{i-1}>}{|| m_i ||.||m_{i-1}||}
\end{aligned}
\end{equation}
where $Dad_i$ represents similarity between two adjacent generated patches, $Dmax_i$ stands for the similarity between $m_i$ and the most similar patch $\overline{m_i}$ in contextual region. Since each generated patch includes the contextual and the previous patch information, $Dad_i$ and $Dmax_i$ are normalized as the weight for the two parts of generated patch. The original patches in $M$ are replaced with generated patches to get a new feature map. We illustrate the process in Fig \ref{img3}.

 \textbf{Search:} We first extract patches in $\overline{M}$ and reshape them as convolutional filters, then apply the convolution filters on $M$. With this operation, we can obtain a vector of values denoting the cross-correlation between each patch in $M$ and all patches in $\overline{M}$. In the end, for generated patch $m_i$, we initialize it with the most similar contextual patch $\overline{m_i}$ and the maximum cross-correlation value $Dmax_{i}$ is recorded for the next step. \textbf{Generate:} The top left patch is taken as the initial patch for the generative process (marked by $m_1$ in Figure 3). Since the $m_1$ has no previous patch, the $Dad_{1}$ is 0 and we replace the $m_1$ with $ \overline {m_1}$ directly, $m_1 = \overline {m_1}$. While the next patch $m_2$ has a previous patch $m_1$ as an additional reference, we therefore view the $m_{1}$ as a convolution filter to measure the cross-correlation metric $Dad_{2}$ between $m_{1}$ and $m_2$. Finally, the $Dad_{2}$ and $Dmax_{2}$ are combined and normalized to the compute of new $m_{2}$, $m_{2}=\frac{Dad_{2}}{Dad_{2}+Dmax_{2}}\times m_{1}+\frac{Dmax_{2}}{Dad_{2}+Dmax_{2}} \times \overline{m_2}$. As mentioned above, from $m_1$ to $m_n$, the generative process can be summarized as:
\begin{equation}
\rm
\begin{gathered}
m_{1}=\overline {m_1}, Dad_1=0 \\
\underset{i\in (2\sim n)}{m_{i}}=\frac{Dad_{i}}{Dad_{i}+Dmax_{i}}\times m_{(i-1)}+\\
\frac{Dmax_{i}}{Dad_{i}+Dmax_{i}} \times \overline {m_i}
\label{eq3}
\end{gathered}
\end{equation}
Since the generate operation is an iterative process, the $m_{i}$ is related to all previous patches($m_{1}$ to $m_{i-1}$ ) and $\overline {m_i}$, each generated patch $m_{i}$ can obtain more contextual information in the meanwhile. We get an attention map $A_{i}$ which records the $\frac{Dmax_{i}}{Dad_{i}+Dmax_{i}}$ and $ \frac{Dad_{i}}{Dad_{i}+Dmax_{i}} \times A_{i-1}$ for $m_{i}$, then $A_{1}$ to $A_{n}$ form a attention matrix, finally the extract patches in $\overline{M}$ are reused as deconvolutional filters to reconstruct $M$. The process of CSA layer is shown in the Algorithm \ref {al1}.

\begin{figure}[t]
\begin{center}
\includegraphics[scale=0.35]{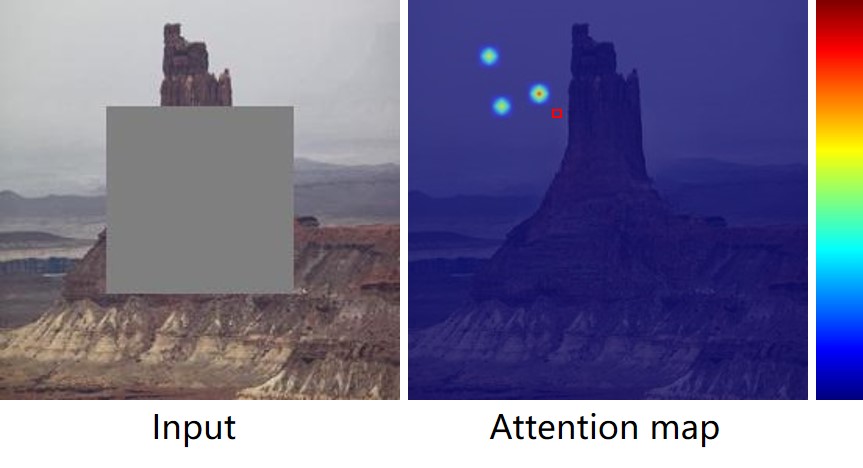}
\end{center}
   \caption{ The visualization of attention map. Dark red means the attention value is large, while light blue means the attention value is small.
}
\label{img4}
\end{figure}

To interpret the CSA layer, we visualize the attention map of a pixel in Fig \ref{img4}, where the red square marks the position of the pixel, the background is our inpainted result, dark red means the attention value is large, while light blue means the attention value is small.

\begin{algorithm}
\caption{Process of CSA layer}
\begin{algorithmic}[1]
\Require The set of feature map for current batch $F_{in}$
\Ensure Reconstructed feature map $F_{out}$
\State \textbf{Search}
\State Reshape $\overline{M}$ as a convolution filter and apply in ${M}$
\State Use Eq (\ref{eq1}) to get the $Dmax_i$ and ${\overline{m_i}}$
\State Initialize ${m_i}$ with ${\overline{m_i}}$
\State \textbf{End search}
\State \textbf{Generate}
\For{$i = 1 \to n$}
\State Use Eq (\ref{eq2}) to calculate the $Dad_{i}$
\State Use Eq (\ref{eq3}) to get the attention map $A_i$ for $m_{i}$
\EndFor
\State Combine $A_{1}$ to $A_{n}$ to get a attention matrix
\State Reuse $\overline{M}$ as a deconvolutional to get F$_{out}$
\State \textbf{End Generate}
\State Return $F_{out}$
\end{algorithmic}
\label {al1}
\end{algorithm}
\subsection{Consistency loss}

Some methods~\cite{6,38} use the perceptual loss~\cite{15} to improve the recognition capacity of the network. However, perceptual loss can not directly optimize the convolutional layer, which may mislead the training process of the CSA layer. Moreover, it does not ensure consistency between the feature maps after the CSA layer and the corresponding layer in the decoder.

We adjust the form of perceptual loss and propose the consistency loss to solve this problem. As shown in Fig \ref{img2}, we use an ImageNet-pretrained VGG-16 to extract a high level feature space in the original image. Next, for any location in $M$, we set the feature space as the target for the CSA layer and the corresponding layer of the CSA in decoder respectively to compute the the $L_2$ distance. In order to match the shape of the feature maps, we adopt $\rm 4_-3$ layer of VGG-16 for our consistency loss. The consistency loss is defined as:
\begin{equation}\label{eq:consistency loss}
\begin{aligned}
L_c = \sum_{y\in M}\lVert CSA( I_{ip})_y-\Phi_{n}(I_{gt})_y \rVert_2^2+\\
\lVert CSA_d(I_{ip})_y-\Phi_{n}(I_{gt})_y \rVert_2^2
\end{aligned}
\end{equation}
Where $\Phi_{n}$  is the activation map of the selected layer in VGG-16. $CSA(.)$ denotes the feature after the CSA layer and $ CSA_d(.)$ is the corresponding feature in the decoder.

Guidance loss is similar to our consistency loss, proposed in~\cite{9}. They view the ground-truth encoder features of the missing parts as a guide to stabilize training. However, extracting the ground truth features by shift-net is an expensive operation, and the semantic understanding ability of shift-net is not as good as VGG network. Moreover, it cannot optimize the specific convolution layer of the encoder and the decoder simultaneously. In summary, our consistency loss fits our requirements better.
\subsection{Feature Patch Discriminator}
\begin{figure}[t]
\begin{center}
\includegraphics[scale=0.26]{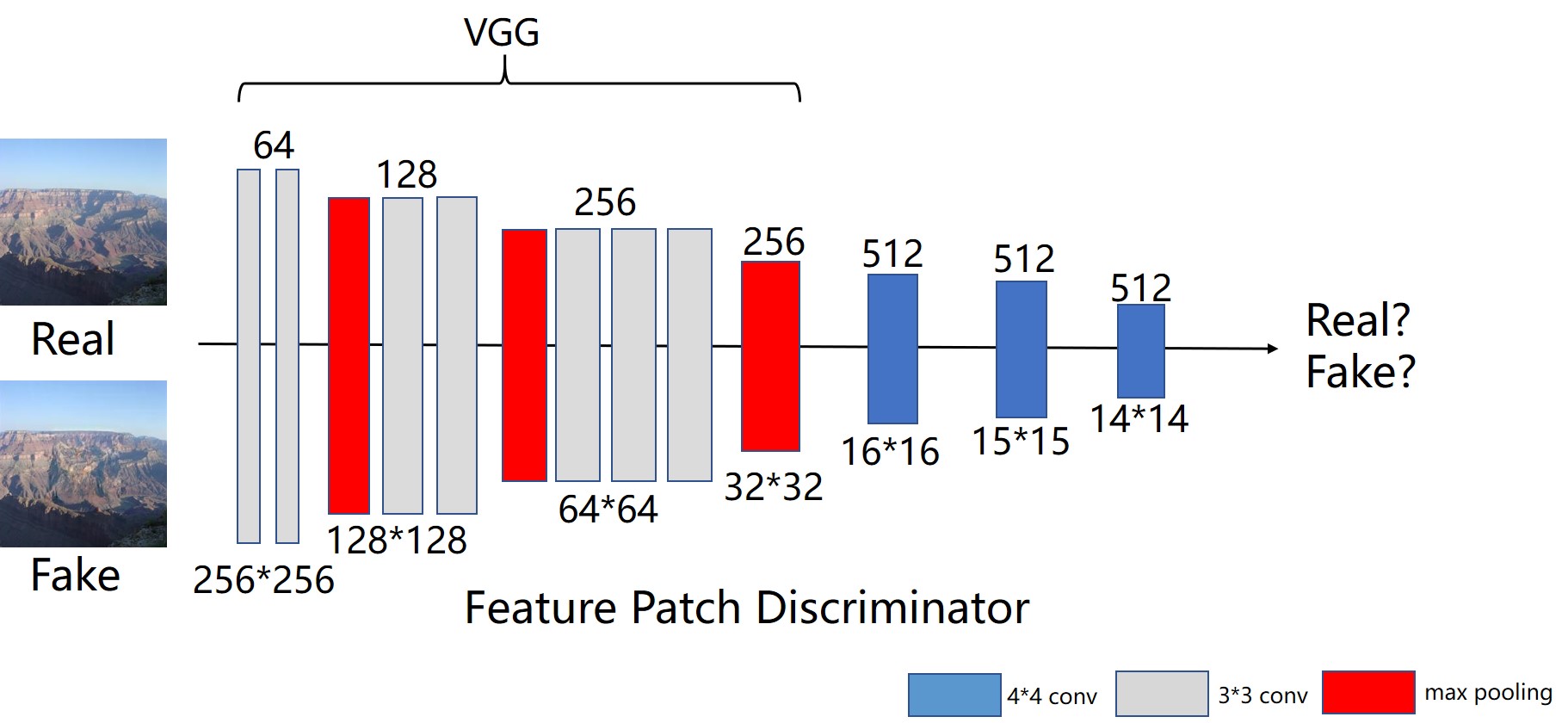}
\end{center}
   \caption{Architecture of our feature patch discriminator network. The number above a convolution layer represents the shape of feature maps.
}
\label{img5}
\end{figure}

Previous image inpainting networks always use an additional local discriminator to improve results. However, the local discriminator is not suitable for irregular holes which may be with any shapes and at any locations. Motivated by Gated Conv~\cite{39}, Markovian Gans~\cite{44} and SRFeat~\cite{16}, we develop a feature patch discriminator to discriminate completed images and original images by inspecting their feature maps. As shown in Fig \ref{img5}, we use VGG-16 to extract feature map after the pool3 layer, then the feature map is treated as an input for several down-sample layers to capture the feature statistics of Markovain patches~\cite{44}. Finally we directly calculate the adversarial loss in this feature map, since receptive fields of each point in this feature map can still cover the entire input image. Our feature patch discriminator combines the advantages of the conventional feature discriminator~\cite{16} and patch discriminator~\cite{11}, which is not only fast and stable during training but also makes the refinement network synthesize more meaningful high-frequency details.

In addition to the feature patch discriminator, we use a 70$\times$70 patch discriminator to discriminate $I_r$ and $I_{gt}$ images by inspecting
their pixel values similar to ~\cite{16}. Meanwhile, we use Relativistic Average LS adversarial loss~\cite{12} for our discriminators. This loss can help refinement network benefit from the gradients from both generated data and real data in adversarial training, which is useful for the training stability. The GAN loss term $D_R$ for refinement network and the loss function $D_F$ for the discriminators are defined as:
\begin{equation}\label{eq:DR}
\begin{aligned}
D_R=-\mathbb{E}_{I_{gt}}[D(I_{gt},I_r)^2]-\mathbb{E}_{I_r}[(1-D(I_r,I_{gt}))^2]
\end{aligned}
\end{equation}
\begin{equation}\label{eq:DF}
\begin{aligned}
D_F=-\mathbb{E}_{I_{gt}}[(1-D(I_{gt},I_r))^2]-\mathbb{E}_{I_r}[D(I_r,I_{gt})^2]
\end{aligned}
\end{equation}
where D stands for the discriminators, $\mathbb{E}_{{I_{gt}/I_f}}$ [.] represents the operation of taking average for all real/fake data in the mini-batch.
\subsection{Objective}
Following the ~\cite{9}, we use $L_1$ distance as our reconstruction loss to make the constrains that the $I_{p}$ and $I_{r}$ should approximate the ground-truth image:
\begin{equation}\label{eq:Reconstruct Loss}
L_{re} = \lVert I_{p}-I_{gt} \rVert_1+\lVert I_{r}-I_{gt} \rVert_1
\end{equation}

Taking consistency, adversarial, and reconstruct losses into account, the overall objective of our refinement network and rough network is defined as:
\begin{equation}\label{eq:final loss}
\begin{aligned}
L = \lambda_r L_{re}+ \lambda_cL_c+\lambda_dD_R
\end{aligned}
\end{equation}
where $\lambda_r$, $\lambda_c$, $\lambda_d$  are the tradeoff parameters for the reconstruction, consistency, and adversarial losses, respectively.

\section{Experiments}
\begin{figure*}[t]
\centering
\includegraphics[scale=0.30]{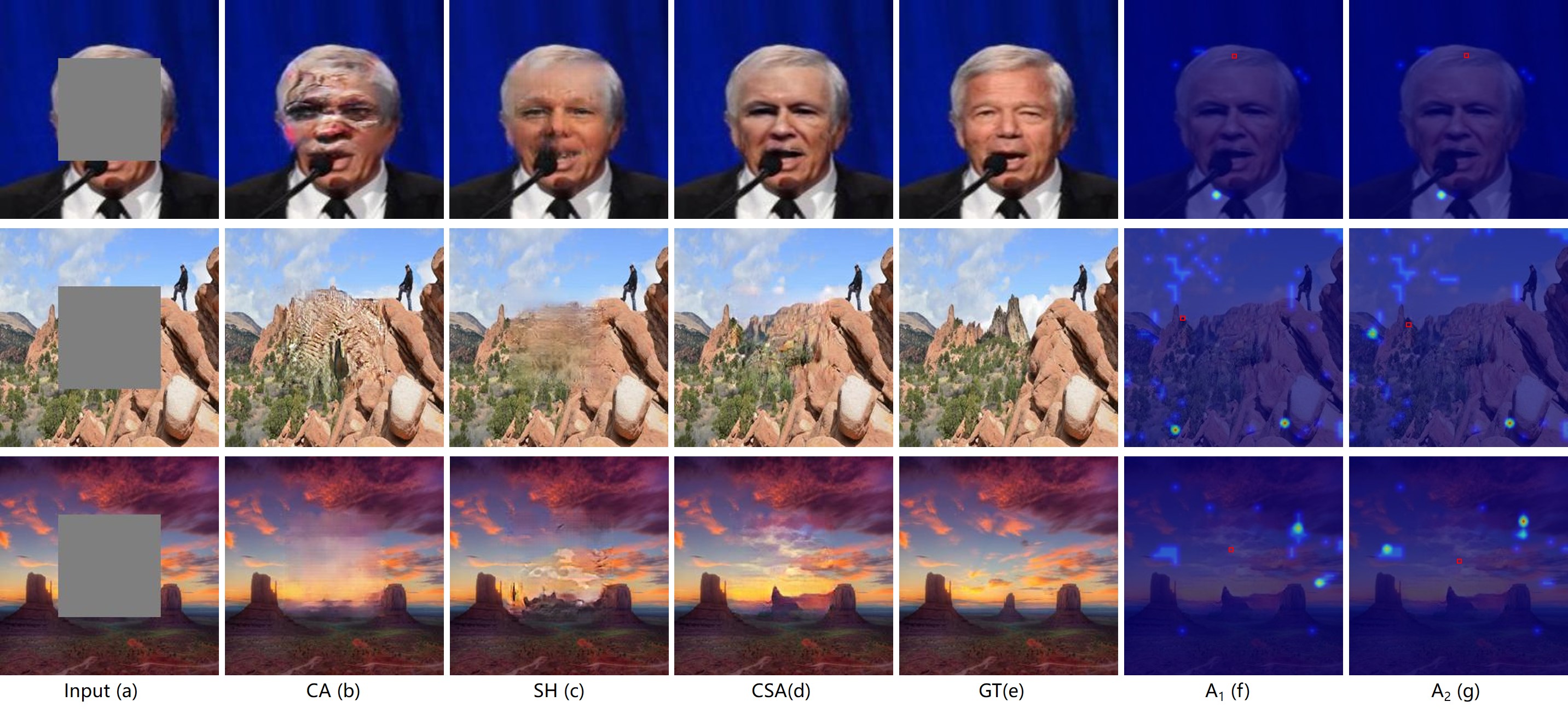}
\caption{Qualitative comparisons in centering masks cases. The first row is the testing result on Celeba image and the others are the testing result on Places2 images.}
\label{img6}
\end{figure*}
\begin{figure*}[t]
\centering
\includegraphics[scale=0.30]{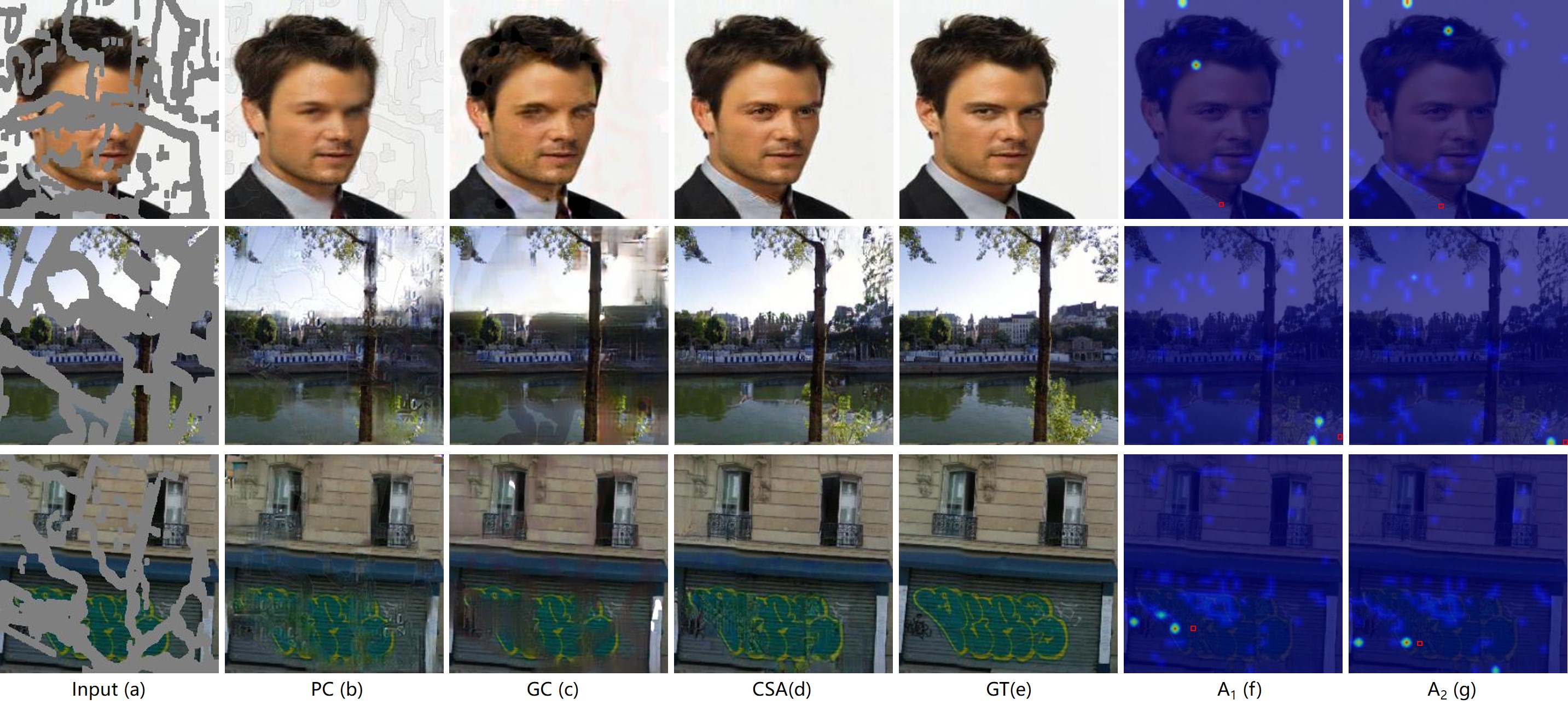}
\caption{ Qualitative comparisons in irregular masks cases. The first row is the testing result on Celeba image and the others are the testing result on Paris StreetView images.}
\label{img7}
\end{figure*}
We evaluate our method on three datasets: Places2~\cite{13}, CelebA~\cite{14}, and Paris StreetView~\cite{40}. We use the original train, test, and validation splits for these three datasets. Data augmentation such as flipping is also adopted during training. Our model is optimized by the Adam algorithm~\cite{42} with a learning rate of $2\times10^{-4}$ and $\beta_1$ = 0.5. The tradeoff parameters are set as $\lambda_r$ =1, $ \lambda_c$=0.01, $\lambda_d$=0.002. We train on a single NVIDIA 1080TI GPU (11GB) with a batch size of 1. The training of CelebA model, Paris StreetView model, Place2 model have taken 9 days, 5 days and 2 days, respectively.

We compare our method with four methods:

-CA: Contextual Attention, proposed by Yu et al. ~\cite{8}

-SH: Shift-net, proposed by Yan et al. ~\cite{9}

-PC: Partial Conv, proposed by Liu et al. ~\cite{38}

-GC: Gated Conv, proposed by Yu et al. ~\cite{39}

 To fairly evaluate, we conduct experiments on both settings of centering and irregular holes. We obtain irregular masks from the work of PC. These masks are classified based on different hole-to-image area ratios (e.g., 0-10(\%), 10-20(\%), etc.). For centering hole, we compare with CA and SH on image from CelebA~\cite{13} and Places2~\cite{14} validation set. For irregular holes, we compare with PC and GC using Paris StreetView~\cite{40} and CelebA~\cite{13} validation images. All the masks and images for training and testing are with the size of 256$\times$256, and our full model runs at 0.82 seconds per frame on GPU for images.
\subsection{Qualitative Comparison }
For centering mask, as shown in Fig \ref{img6}, CA~\cite{8} is effective in semantic inpainting, but the results present distorted structure and confusing color. SH~\cite{9} performances better due to the shift operation and guidance loss, but its predictions are to some extent blurry and detail-missing. For irregular mask, as shown in Fig \ref{img7}, PC~\cite{38} and GC~\cite{39} can get smooth and plausible result, but the continuities in color and lines do not hold well and some artifacts can still be observed on generated images. This is mainly due to the fact that these methods do not consider the correlations between the deep features in hole regions. In comparison to these competing methods, our model can handle these problems better, and generate visually pleasing results. Moreover, as shown in Fig \ref{img6} and Fig \ref{img7} (f, g), A$_1$ and A$_2$ are attention maps of two adjacent pixels, the first line is the attention maps of left and right adjacent pixels, the second and third line is the attention maps of up and down adjacent pixels. We see that the attention maps of two adjacent pixels are basically the same, and the perceived areas are not limited to the most relevant contextual areas. These phenomena can prove that our approach is better at modeling the coherence of the generated content and enlarging the perception domain for each generated patch than other attention based model~\cite{8,9}.
\subsection{Quantitative comparisons }

We randomly select 500 images from Celeba validation dataset~\cite{13} and generate irregular and centering holes for each image to make comparisons. Following the CA~\cite{8}, we use common evaluation metrics, i.e., L1, L2, PSNR, and SSIM to quantify the performance of the models. Table \ref{tab1} and Table \ref{tab2} list the evaluation results with centering mask and irregular masks respectively. It can be seen that our method outperforms all the other methods on these measurements with irregular mask or centering mask.
\begin{table}[t]

\centering
 \begin{tabular}{|l|c|c|c|c|}
 \hline

    & L$_1^-$(\%)      & L$_2^-$(\%)    & SSIM$^+$          & PSNR$^+$              \\ \hline
 CA         & 2.64          & 0.47          & 0.882          & 23.93               \\
 SH        & 1.97         & 0.28          & 0.926       & 26.38               \\\hline
 CSA         & \textbf{1.83}         & \textbf{0.27}            & \textbf{0.931}         & \textbf{26.54}                 \\\hline
 \end{tabular}

 \caption{Comparison results over Celeba with centering hole between CA~\cite{8}, SH~\cite{9}, and Ours. $^-$Lower is better. $^+$Higher is better}
\label{tab1}
\end{table}
\begin{table}[t]
\centering
 \begin{tabular}{|l|c|c|c|c|c|}
 \hline

         & Mask         & PC         & GC         & CSA     \\ \hline
          &10-20\%      & 1.00        & 1.00           &\textbf{0.72}   \\
L$_1^-$(\%) &20-30\%    & 1.46       & 1.40           & \textbf{0.94} \\
         &30-40\%       & 2.97       & 2.62           & \textbf{2.18}     \\
          &40-50\%      & 4.01      & 3.26           & \textbf{2.85}     \\\hline
          &10-20\%      & 0.12       & 0.08           &\textbf{0.04}    \\
L$_2^-$(\%)&20-30\%     & 0.19       & 0.12           & \textbf{0.07} \\
         &30-40\%       & 0.58       & 0.44           & \textbf{0.37}     \\
          &40-50\%      & 0.76       & 0.50           & \textbf{0.44}      \\\hline

          &10-20\%      & 31.13     & 31.67         &\textbf{34.69}    \\
PSNR$^+$  &20-30\%      & 29.10      & 29.83          & \textbf{32.58} \\
         &30-40\%       & 23.46     &24.48           & \textbf{25.32}     \\
          &40-50\%      & 22.11     & 23.36          & \textbf{24.14}      \\\hline

          &10-20\%      & 0.970        & 0.977           &\textbf{0.989}   \\
SSIM$^+$  &20-30\%      & 0.956        & 0.964           & \textbf{0.982} \\
         &30-40\%       & 0.897       & 0.910           & \textbf{0.926}     \\
          &40-50\%      & 0.839      & 0.860          & \textbf{0.883}      \\\hline

 \end{tabular}
 \caption{Comparison results over Celeba with irregular mask between PC~\cite{38}, GC~\cite{39}, and Ours. $^-$Lower is better. $^+$Higher is better}
\label{tab2}
\end{table}
\subsection{Ablation Study}

\begin{figure}[h]
\centering
\includegraphics[scale=0.20]{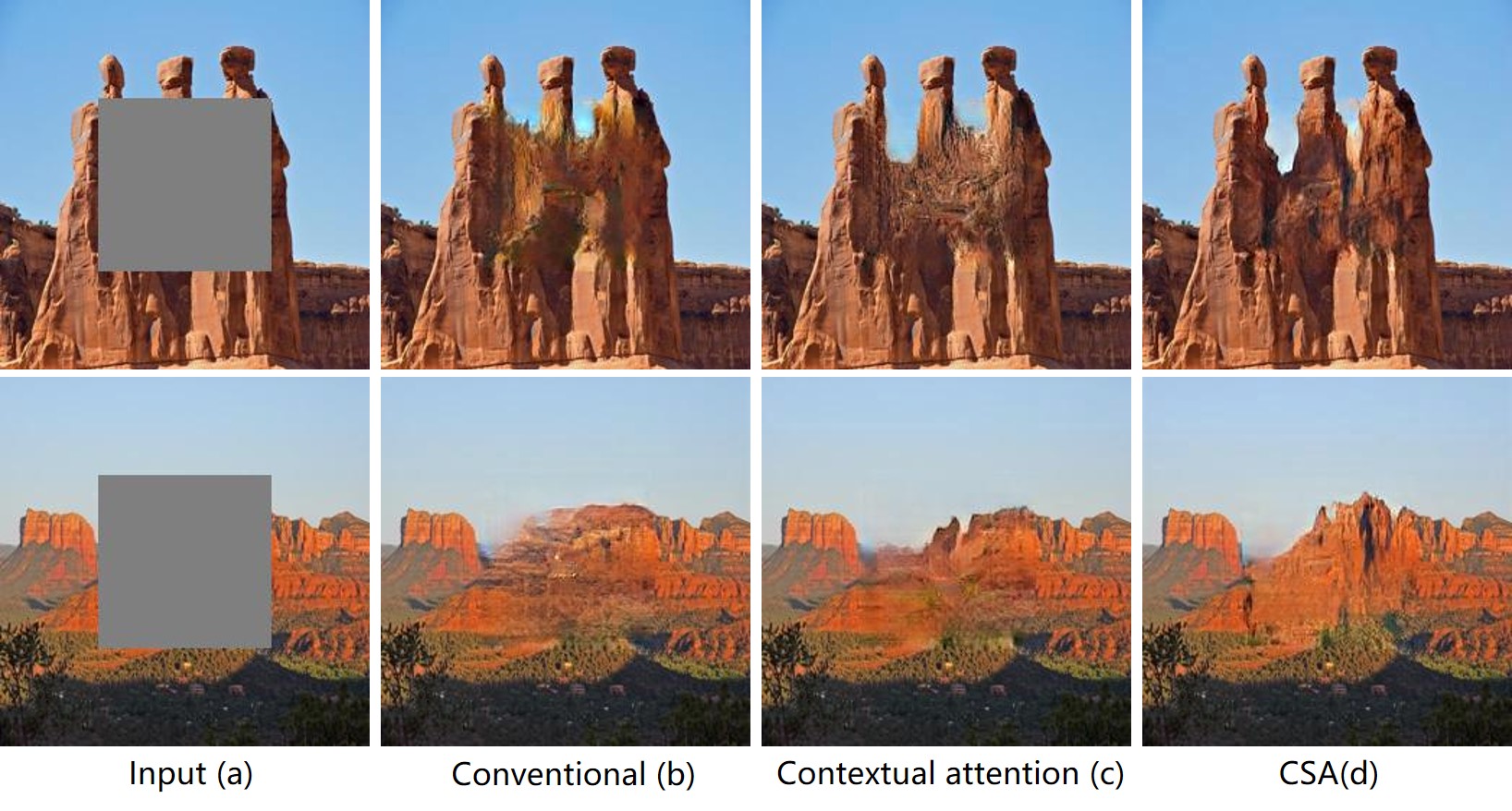}
\caption{The effect of CSA layer. (b), (c) are results of our model which replace the CSA layer with the conventional layer and the CA layer~\cite{8} respectively.}
\label{img8}
\end{figure}

\textbf{Effect of CSA layer}
To investigate the effectiveness of CSA, we replace the CSA layer with a conventional 3$\times $3 layer and the contextual attention layer~\cite{8} respectively to make a comparison. As shown in Fig \ref{img8}(b), the mask part fails to restore reasonable content when we use conventional conv. Although contextual attention layer~\cite{8} can improve the performance compared to conventional convolution, the inpainting results still lack fine texture details and the pixels are not consistent with the background(see Fig \ref{img8}(c)). Compared with them, our method performs better (see Fig \ref{img8}(d)). This illustrates the fact that the global semantic structure and local coherency are constructed by the CSA layer.

\begin{figure}[h]
\centering
\includegraphics[scale=0.20]{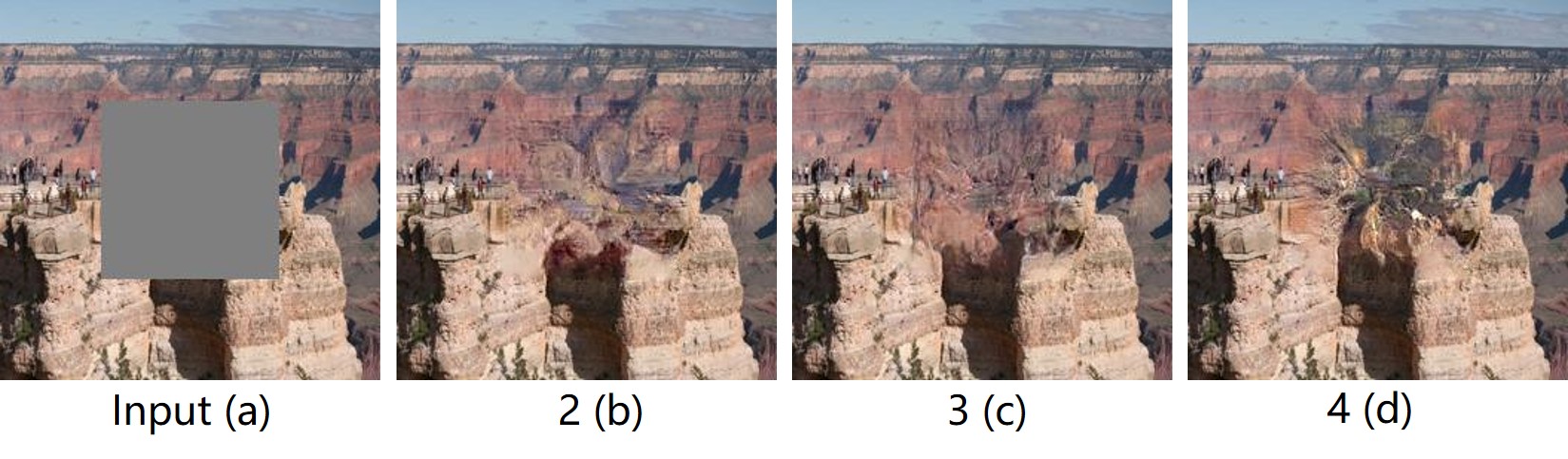}
\caption{The results of CSA layer on three down-sample positions of refinement network: 2nd, 3rd, and 4th.}
\label{img9}
\end{figure}

\textbf{Effect of CSA layer at different positions}
Too deep or too shallow positions of CSA layer may cause loss of information details or increase calculation time overhead. Fig \ref{img9} shows the results of the CSA layer at the 2nd, 3rd, and 4th down-sample positions of refinement network. When the CSA layer is placed on the 2nd position with 64$\times$64 size (See Fig \ref{img9}(b)), our model performances well but it takes more time to process an image. When the CSA layer is placed on 4th position with 16$\times$16 size (See Fig \ref{img9}(c)), our model becomes very efficient but tends to generate the result with coarse details. By performing the CSA layer in the 3rd position with 32$\times$32 size, better tradeoff between efficiency (i.e., 0.82 seconds per image) and performance can be obtained by our model (See Fig \ref{img9}(d)).

\begin{figure}[h]
\centering
\includegraphics[scale=0.20]{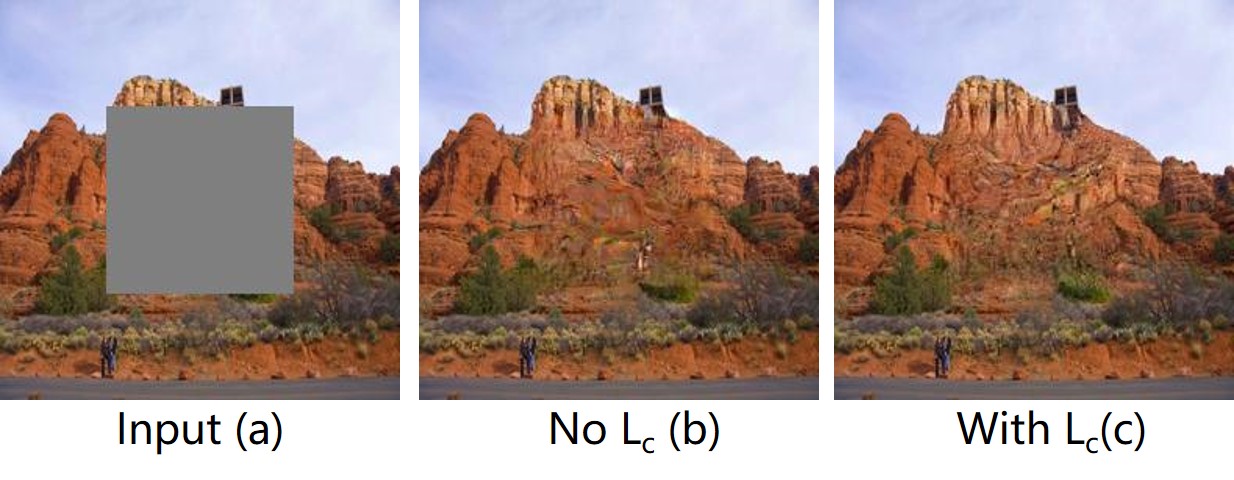}
\caption{The effect of consistency loss. (b), (c) are results of our model without or with consistency loss}
\label{img10}
\end{figure}

\textbf{Effect of consistency loss}
We conduct further experiment to evaluate the effect of consistency loss. We add and drop out the consistency loss $\rm L_c$ to train the inpainting model. Fig \ref{img10} shows the comparison results. It can be seen that, without the consistency loss, the center of the hole regions present distorted structure, which may be due to training instability and misunderstanding of image semantic [See Fig \ref{img10}(b)]. The consistency loss helps to deal with these issues [See Fig \ref{img10}(c)].

\begin{figure}[h]
\centering
\includegraphics[scale=0.20]{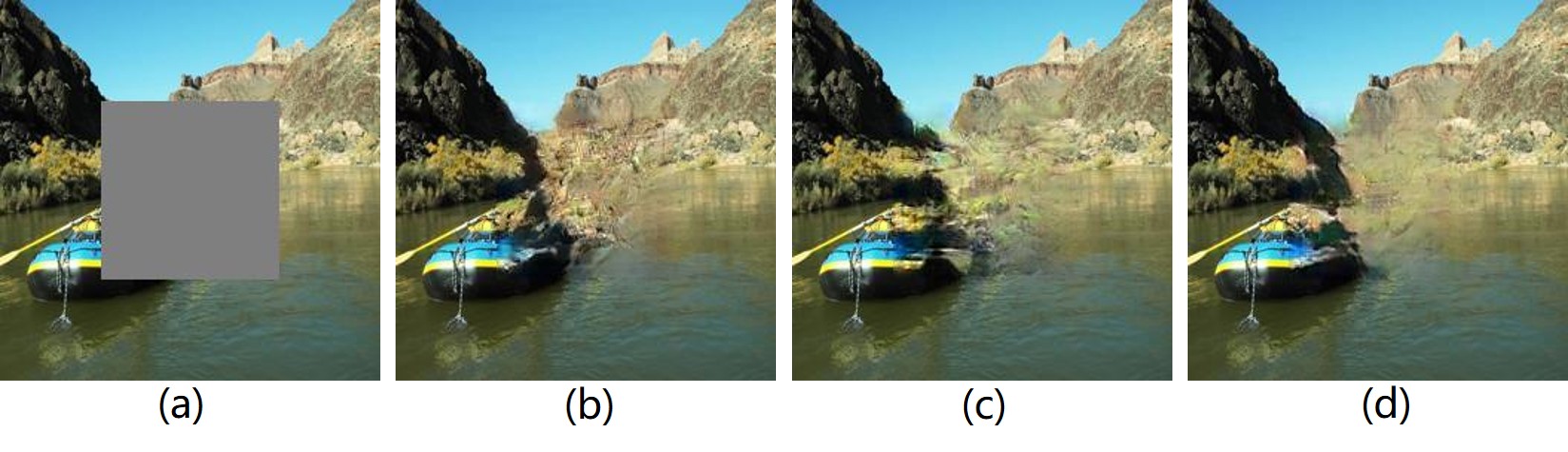}
\caption{The effect of feature patch discriminator. Given the input (a), (b), (c) and (d) are the results when we use patch discriminator, patch and SRFeat feature discriminators ~\cite{16}, patch and feature patch discriminators, respectively.
}
\label{img11}
\end{figure}

\textbf{Effect of feature patch discriminator }
 As shown in Fig \ref{img11}(b), when we only use the patch discriminator, the result performances distorted structure. Then we add the conventional feature discriminator~\cite{16}, however the generated content still seems blurry (See Fig \ref{img11}(c)). Finally, by performing the feature patch discriminator, fine details and reasonable structure can be obtained (See Fig \ref{img11}(d)). Moreover, the feature patch discriminator processes each image for 0.2 seconds faster than the conventional one~\cite{16}.
\vspace{-0.2cm}
\section{Conclusion}
In this paper, we proposed a fined deep generative model based approach which designed a novel Coherent Semantic Attention layer to learn the relationship between features of missing region in image inpainting task. The consistency loss is introduced to enhance the CSA layer learning ability for ground truth feature distribution and training stability. Moreover, a feature patch discriminator is joined into our model to achieve better predictions. Experiments have verified the effectiveness of our proposed methods. In future, we plan to extend the method to other tasks, such as style transfer and single image super-resolution.

{\small
\bibliographystyle{ieee}
\bibliography{egpaper_for_review}
}

\setcounter{section}{0}
\renewcommand\thesection{\Alph{section}}
\section{Definition of Masked Region in Feature Maps}
As the CSA layer works based on both the masked region $M$ and unmasked region $\overline{M}$ in feature maps, thus we need to give a definition of masked region in feature maps. In our implementation, we introduce a masked image in which each pixel value of known regions is 0 and that for unknown regions is 1. When considering centering masks, since the CSA layer locates at the resolution of 32$\times$32 and the centering mask covers half of the input image I$_{in}$, we set the size of region $M$ in feature maps as 16$\times$16. While for irregular masks, following the idea of SH ~\cite{9}, we first define a network that has the same architecture with the encoder of rough network but with the network width of 1, the network has only convolution layers and all the elements of the filters are 1/16. Then taking the masked image as input, we obtain the feature with 32$\times$32 resolution which is the 3rd down-sample output of the network. Finally, for the value at each position of the feature, we set those values larger than 5/16 to 1, which means this position belongs to masked region $M$ in feature maps.
\section{Network Architectures }
As a supplement to the content of Section 3, we will report more details of our network architectures in the following. First, Table \ref{taba1} and Table \ref{taba4} depict the specific design of architecture of our rough network and refinement network respectively. On one hand, the architecture of rough network is the same as pix to pix [19]. On the other hand, the refinement network uses 3$\times$3 convolutions to double the channel and uses 4$\times$4 convolutions to reduce the spatial size to half. Then, the architecture of patch and feature patch discriminators are shown in Table \ref{taba2} and Table \ref{taba3} respectively, where the VGG 4-3 denotes all the layers before Relu 4$_-$3 of VGG-16 network.
\begin{table}[ht]

\centering
 \begin{tabular}{|l|c|c|c|c|}
 \hline

 \textbf{The architecture of rough network }          \\ \hline
  [Layer 1] Conv. (4, 4, 64), stride=2;            \\\hline
  [Layer 2] LReLU; Conv. (4, 4, 128), stride=2; IN;       \\\hline
  [Layer 3] LReLU; Conv. (4, 4, 256), stride=2; IN;        \\\hline
  [Layer 4] LReLU; Conv. (4, 4, 512), stride=2; IN;        \\\hline
  [Layer 5] LReLU; Conv. (4, 4, 512), stride=2; IN;        \\\hline
  [Layer 6] LReLU; Conv. (4, 4, 512), stride=2; IN;        \\\hline
  [Layer 7] LReLU; Conv. (4, 4, 512), stride=2; IN;        \\\hline
  [Layer 8] LReLU; Conv. (4, 4, 512), stride=2;        \\\hline
  [Layer 9] ReLU; DeConv. (4, 4, 512), stride=2; IN;     \\
                             Concatenate(Layer 9, Layer 7);       \\\hline
  [Layer 10] ReLU; DeConv. (4, 4, 512), stride=2; IN;       \\
  Concatenate(Layer 10, Layer 6);       \\\hline
  [Layer 11] ReLU; DeConv. (4, 4, 512), stride=2; IN;   \\
  Concatenate(Layer 11, Layer 5);       \\\hline
  [Layer 12] ReLU; DeConv. (4, 4, 512), stride=2; IN;  \\
  Concatenate(Layer 12, Layer 4);       \\\hline
  [Layer 13] ReLU; DeConv. (4, 4, 256), stride=2; IN;  \\
  Concatenate(Layer 13, Layer 3);       \\\hline
  [Layer 14] ReLU; DeConv. (4, 4, 128), stride=2; IN;  \\
  Concatenate(Layer 16, Layer 2);       \\\hline
  [Layer 15] ReLU; DeConv. (4, 4, 64), stride=2; IN;       \\
  Concatenate(Layer 17, Layer 1);       \\\hline
  [Layer 16] ReLU; DeConv. (4, 4, 3), stride=2; Tanh;         \\\hline
 \end{tabular}

 \caption{The architecture of the Rough network. IN represents InstanceNorm and LReLU donates leaky ReLU with the slope of
0.2.}
\label{taba1}
\end{table}
\begin{table}[ht]

\centering
 \begin{tabular}{|l|c|c|c|c|}
 \hline

 \textbf{The architecture of patch discriminator}          \\ \hline
  [layer 1] Conv. (4, 4, 64), stride=2; LReLU;            \\\hline
  [layer 2] Conv. (4, 4, 128), stride=2; IN; LReLU;   \\\hline
  [layer 3] Conv. (4, 4, 256), stride=2; IN; LReLU;  \\\hline
  [layer 4] Conv. (4, 4, 512), stride=1; IN; LReLU;  \\\hline
  [layer 5] Conv. (4, 4, 1), stride=1;      \\\hline
 \end{tabular}

 \caption{The architecture of the patch discriminative network. ¡°IN¡± represents InstanceNorm and ¡°LReLU¡± donates leaky ReLU with the slope of 0.2.
}
\label{taba2}
\end{table}

\begin{table}[ht]

\centering
 \begin{tabular}{|l|c|c|c|c|}
 \hline

 \textbf{The architecture of feature patch discriminator}          \\ \hline
  [layer 1] VGG 4$_-$3 layer \\\hline
  [layer 2] Conv. (4, 4, 512), stride=2; LReLU;            \\\hline
  [layer 3] Conv. (4, 4, 512), stride=1; IN; LReLU;   \\\hline
  [layer 4] Conv. (4, 4, 512), stride=1;      \\\hline
 \end{tabular}

 \caption{The architecture of the feature patch discriminative network. ¡°IN¡± represents InstanceNorm and ¡°LReLU¡± donates leaky ReLU with the slope of 0.2.
}
\label{taba3}
\end{table}
\section{Quantitative Comparison of Ablation Study}
\textbf{Effect of CSA layer} When examining the effect of CSA layer, we select validation images from $butte$ categories of Places2 dataset and replace the CSA layer with a conventional 3$\times$3 layer and the contextual attention layer~\cite{8} respectively. Table \ref{taba5} lists the evaluation results. From the results in Table \ref{taba5}, we can see that the CSA layer outperforms all the other layers.
\begin{table}[h]
\centering
 \begin{tabular}{|l|c|c|c|c|}
 \hline

                   & L$_1^-$(\%)      & L$_2^-$(\%)     & SSIM$^+$          & PSNR$^+$              \\ \hline
 With Conv         & 2.56            & 0.54            & 0.819             & 23.71               \\
 With CA           & 2.51             & 0.56            & 0.817             & 23.74               \\\hline
 With CSA          & \textbf{2.37}    & \textbf{0.52}    & \textbf{0.823}    & \textbf{24.04}                 \\\hline
 \end{tabular}

 \caption{The effect of CSA layer. $^-$Lower is better. $^+$Higher is better}
\label{taba5}
\end{table}

\textbf{Effect of CSA layer at different positions} In order to compare the effect of CSA layer at different positions, we select validation images from $canyon$ categories of Places2 dataset to make quantitative comparisons. Table \ref{taba6} lists the evaluation results. From the results in Table \ref{taba6}, we find that better tradeoff between efficiency and performance can be achieved by our model when the CSA layer is embedded into the 3th down-sample positions.

\begin{table}[h]
\centering
 \begin{tabular}{|l|c|c|c|c|}
 \hline

                   & L$_1^-$(\%)      & L$_2^-$(\%)     & SSIM$^+$          & PSNR$^+$              \\ \hline
4                  & 3.06            & 0.75            & 0.797             & 22.14               \\
2                 & 2.92            & \textbf{0.70}                   & \textbf{0.803 }            & \textbf{22.61 }              \\\hline
3                  & \textbf{2.83}    & 0.71          & 0.802             & 22.48     \\\hline

 \end{tabular}

 \caption{The effect of CSA layer at different positions. $^-$Lower is better. $^+$Higher is better}
\label{taba6}
\end{table}

\textbf{Effect of consistency loss} In order to verify the validity of consistency loss $L_c$ , we select validation images from $butte$ categories of Places2 dataset to make quantitative comparisons. Table \ref{taba7} lists the evaluation results. From the results in Table \ref{taba7}, we can see that the consistency loss can help our model performances better.
\begin{table}[h]
\centering
 \begin{tabular}{|l|c|c|c|c|}
 \hline

                   & L$_1^-$(\%)      & L$_2^-$(\%)     & SSIM$^+$          & PSNR$^+$              \\ \hline
No $L_c$                  & 2.39            & 0.53           & 0.823             & 23.92               \\\hline
With $L_c$           & \textbf{2.37}    & \textbf{0.52}    & \textbf{0.823}    & \textbf{24.04}     \\\hline

 \end{tabular}

 \caption{The effect of consistency loss. $^-$Lower is better. $^+$Higher is better}
\label{taba7}
\end{table}

\textbf{Effect of feature patch discriminator} We further conduct experiments to validate the effect of feature patch discriminator. We select validation images from $canyon$ categories of Places2 dataset to make quantitative comparisons. Table \ref{taba8} lists the evaluation results. From the results in Table \ref{taba8}, it can be seen that our feature patch discriminator is better than others.
\begin{table}[h]
\centering
 \begin{tabular}{|l|c|c|c|c|}
 \hline

                   & L$_1^-$(\%)      & L$_2^-$(\%)     & SSIM$^+$          & PSNR$^+$              \\ \hline
a                  & 3.07            & 0.77           & 0.793            & 22.12              \\
b                  & 2.99            & 0.77           & 0.794            & 22.16    \\\hline
c                  & \textbf{2.83}    & \textbf{0.71}    & \textbf{0.802}    & \textbf{22.48}     \\\hline
 \end{tabular}

 \caption{The effect of feature patch discriminator. a, b and c are respectively the results when we use patch discriminator,patch and SRFeat feature discriminators [29], patch and our feature patch discriminators. $^-$Lower is better. $^+$Higher is better}
\label{taba8}
\end{table}
\section{More Comparisons Results}
\begin{figure*}[t]
\centering
\includegraphics[scale=0.35]{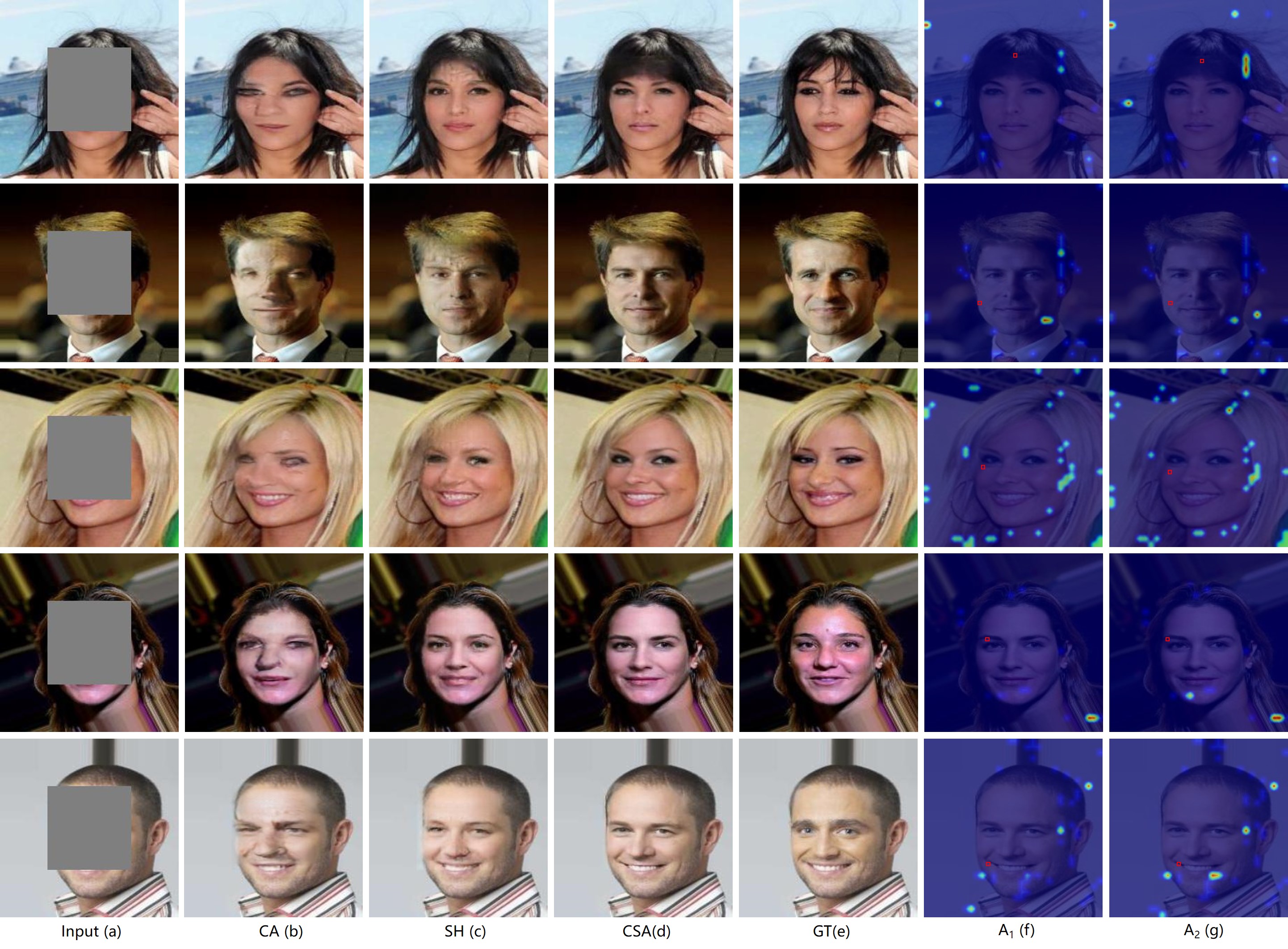}
\caption{Qualitative comparisons on Celeba with centering masks. A$_1$ and A$_2$  are attention maps of two adjacent pixels, the 1st, 2nd, and 3rd rows are the attention maps of up and down adjacent pixels, the 4th and 5th rows are the attention maps of left and right adjacent pixels.}
\label{imga1}
\end{figure*}

\begin{figure*}[t]
\centering
\includegraphics[scale=0.35]{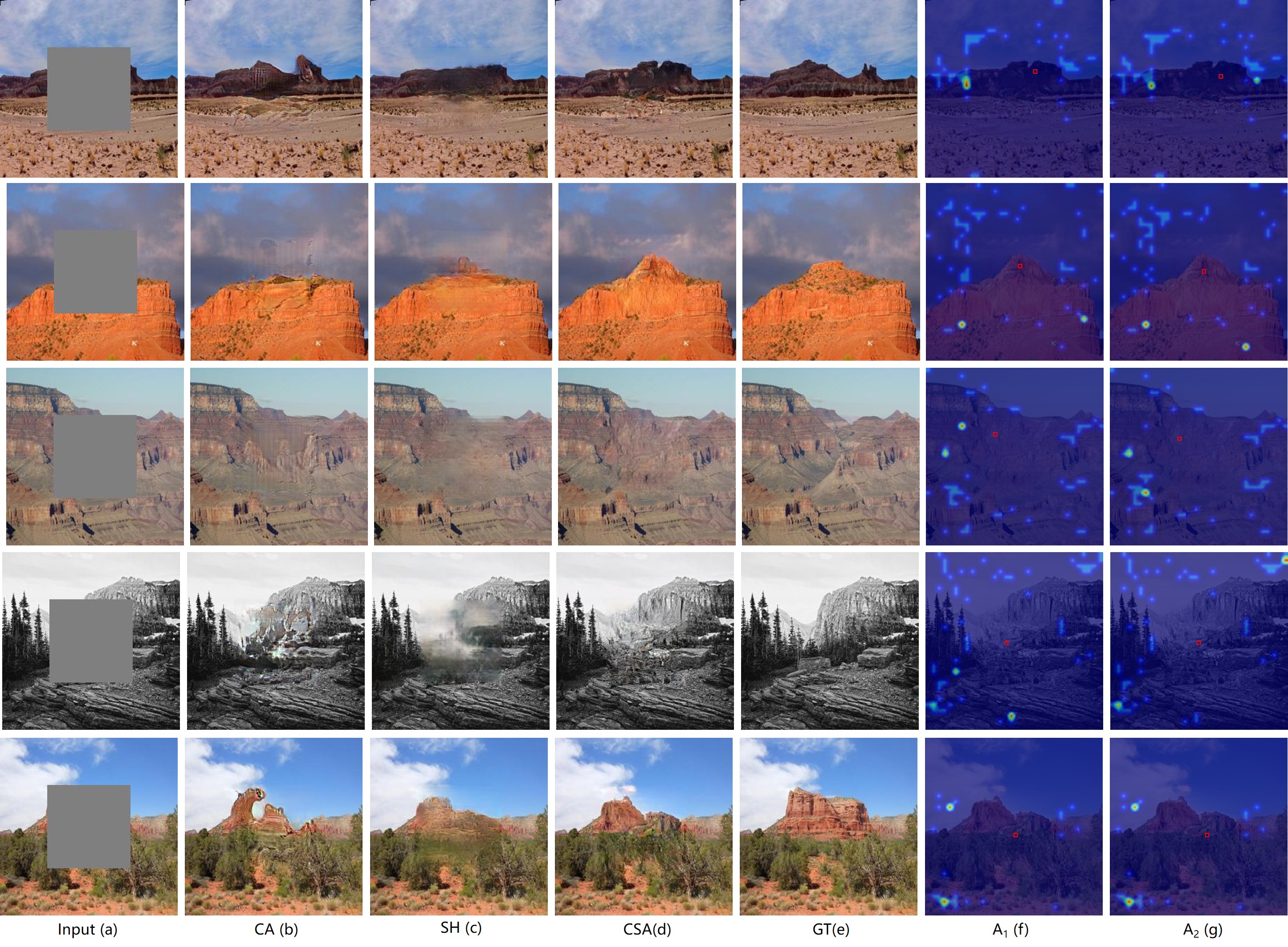}
\caption{Qualitative comparisons on Place2 with centering masks. A$_1$  and A$_2$  are attention maps of two adjacent pixels, the 1st, 2nd, and 3rd rows are the attention maps of up and down adjacent pixels, the 4th and 5th rows are the attention maps of left and right adjacent pixels.}
\label{imga2}
\end{figure*}

\begin{figure*}[t]
\centering
\includegraphics[scale=0.35]{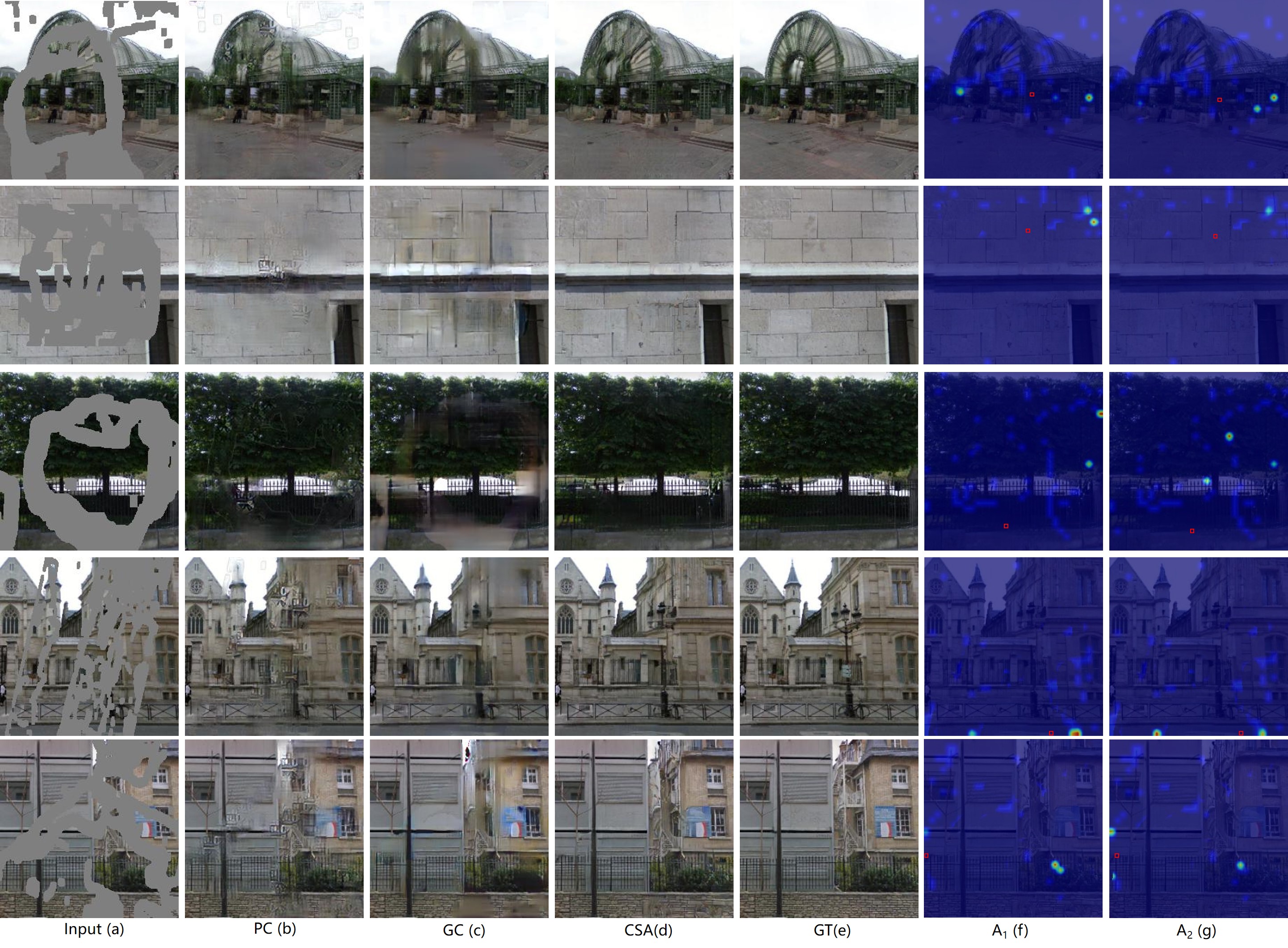}
\caption{ Qualitative comparisons on Paris StreetView with irregular masks. A$_1$ and A$_2$  are attention maps of two adjacent pixels, the 1st, 2nd, and 3rd rows are the attention maps of up and down adjacent pixels, the 4th and 5th rows are the attention maps of left and right adjacent pixels.}
\label{imga3}
\end{figure*}

\begin{figure*}[t]
\centering
\includegraphics[scale=0.35]{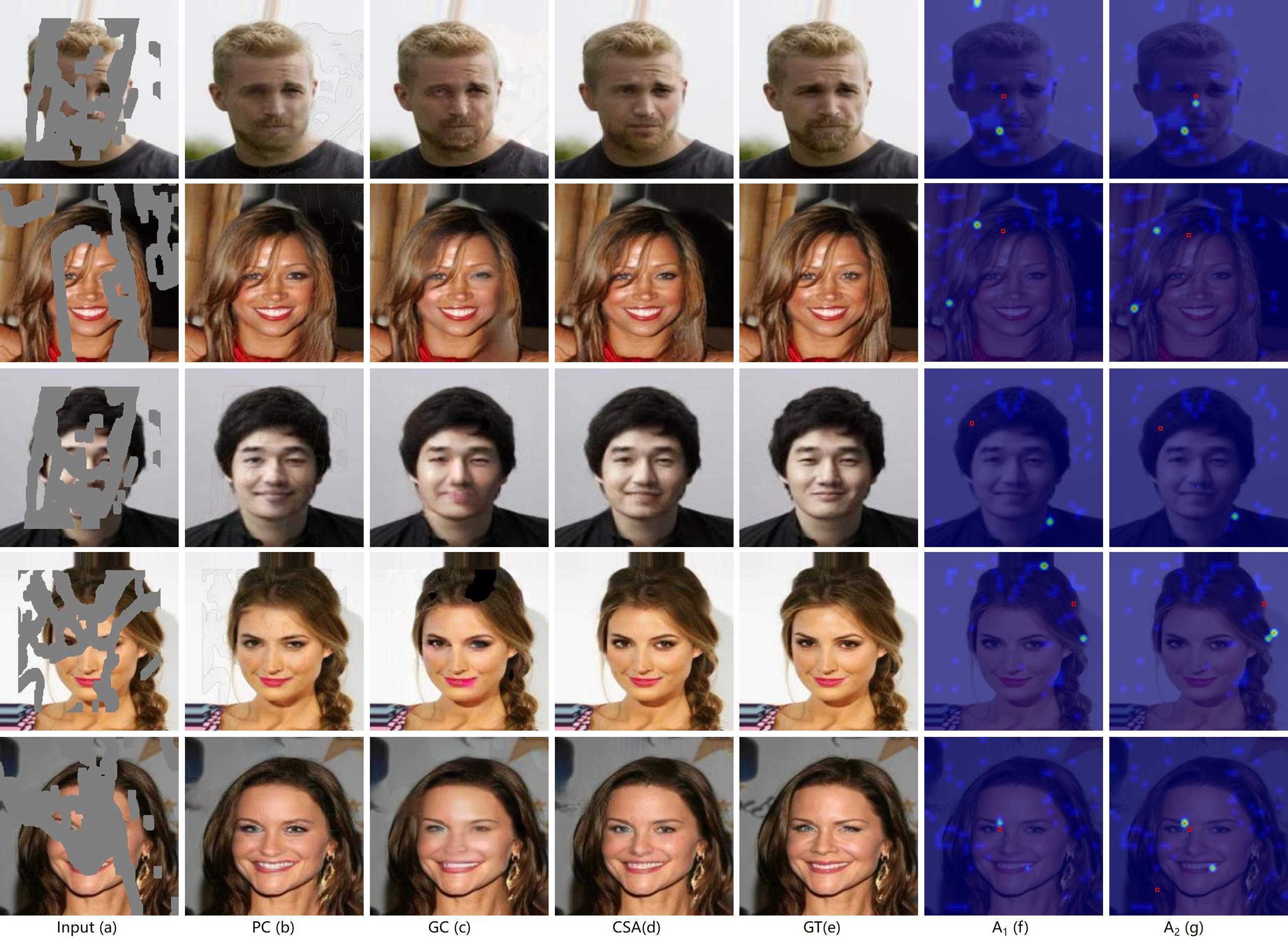}
\caption{Qualitative comparisons on CelebA with irregular masks. A$_1$ and A$_2$ are attention maps of two adjacent pixels, the 1st, 2nd, and 3rd rows are the attention maps of up and down adjacent pixels, the 4th and 5th rows are the attention maps of left and right adjacent pixels}
\label{imga4}
\end{figure*}

More comparisons with CA~\cite{8}, SH~\cite{9}, PC~\cite{38} and GC~\cite{39} on Paris StreetView~\cite{40}, Places2~\cite{14} and CelebA~\cite{13} are also conducted. Please refer to Fig \ref{imga1} and \ref{imga2} for more results on Places2 and CelebA with centering mask. And for comparison on irregular masks, please refer to Fig \ref{imga3} and \ref{imga4} for results on Paris StreetView and CelebA datasets. Table \ref{taba9} lists the evaluation results with centering mask on Place2 dataset, the scene categories selected from Places2 is $butte$. Table \ref{taba10} lists the evaluation results with irregular masks on Paris StreetView dataset. It is obvious that our model outperforms state-of-the-art approaches in both structural consistency and detail richness, and the local pixel continuity is well assured since the CSA layer considers the semantic relevance between the holes features. As a side contribution, we will release the pre-trained model and codes.
\begin{table}[h]
\centering
 \begin{tabular}{|l|c|c|c|c|}
 \hline

                   & L$_1^-$(\%)      & L$_2^-$(\%)     & SSIM$^+$          & PSNR$^+$              \\ \hline
CA                 & 4.08            & 1.02           & 0.704            & 20.69              \\
SH                  & 4.04            & 0.91           & 0.738           & 21.55    \\\hline
CSA                  & \textbf{2.37}    & \textbf{0.52}    & \textbf{0.823}    & \textbf{24.04}     \\\hline
 \end{tabular}

 \caption{Comparison results over Place2 ($butte$) with centering hole between CA [42], SH [36], and Ours. $^-$Lower is better. $^+$Higher is better}
\label{taba9}
\end{table}
\begin{table}[h]
\centering
 \begin{tabular}{|l|c|c|c|c|c|}
 \hline

         & Mask         & PC         & GC         & CSA     \\ \hline
          &10-20\%      & 1.47        & 1.14           &\textbf{1.05}   \\
L$_1^-$(\%) &20-30\%    & 2.12       & 1.71           & \textbf{1.41} \\
         &30-40\%       & 3.49       & 3.19           & \textbf{2.69}     \\
          &40-50\%      & 4.58      & 4.49           & \textbf{3.70}     \\\hline
          &10-20\%      & 0.17       & 0.14           &\textbf{0.08}    \\
L$_2^-$(\%)&20-30\%     & 0.28       & 0.22           & \textbf{0.13} \\
         &30-40\%       & 0.60       & 0.57           & \textbf{0.45}     \\
          &40-50\%      & 0.86       & 0.90           & \textbf{0.68}      \\\hline

          &10-20\%      & 28.91     & 29.58         &\textbf{32.67}    \\
PSNR$^+$  &20-30\%      & 26.78      & 27.43          & \textbf{30.32} \\
         &30-40\%       & 23.27     &23.19           & \textbf{24.85}     \\
          &40-50\%      & 21.67     & 21.33          & \textbf{23.10}      \\\hline

          &10-20\%      & 0.937        & 0.945           &\textbf{0.972}   \\
SSIM$^+$  &20-30\%      & 0.894        & 0.920           & \textbf{0.951} \\
         &30-40\%       & 0.815       & 0.846           & \textbf{0.873}     \\
          &40-50\%      & 0.678      & 0.731          & \textbf{0.768}      \\\hline

 \end{tabular}
 \caption{Comparison results over Paris StreetView  with irregular mask between PC [32], GC [19], and Ours. $^-$Lower is better. $^+$Higher is better}
\label{taba10}
\end{table}
\section{More Results on CelebA, Paris StreetView, Places2}
\begin{figure*}[t]
\centering
\includegraphics[scale=0.6]{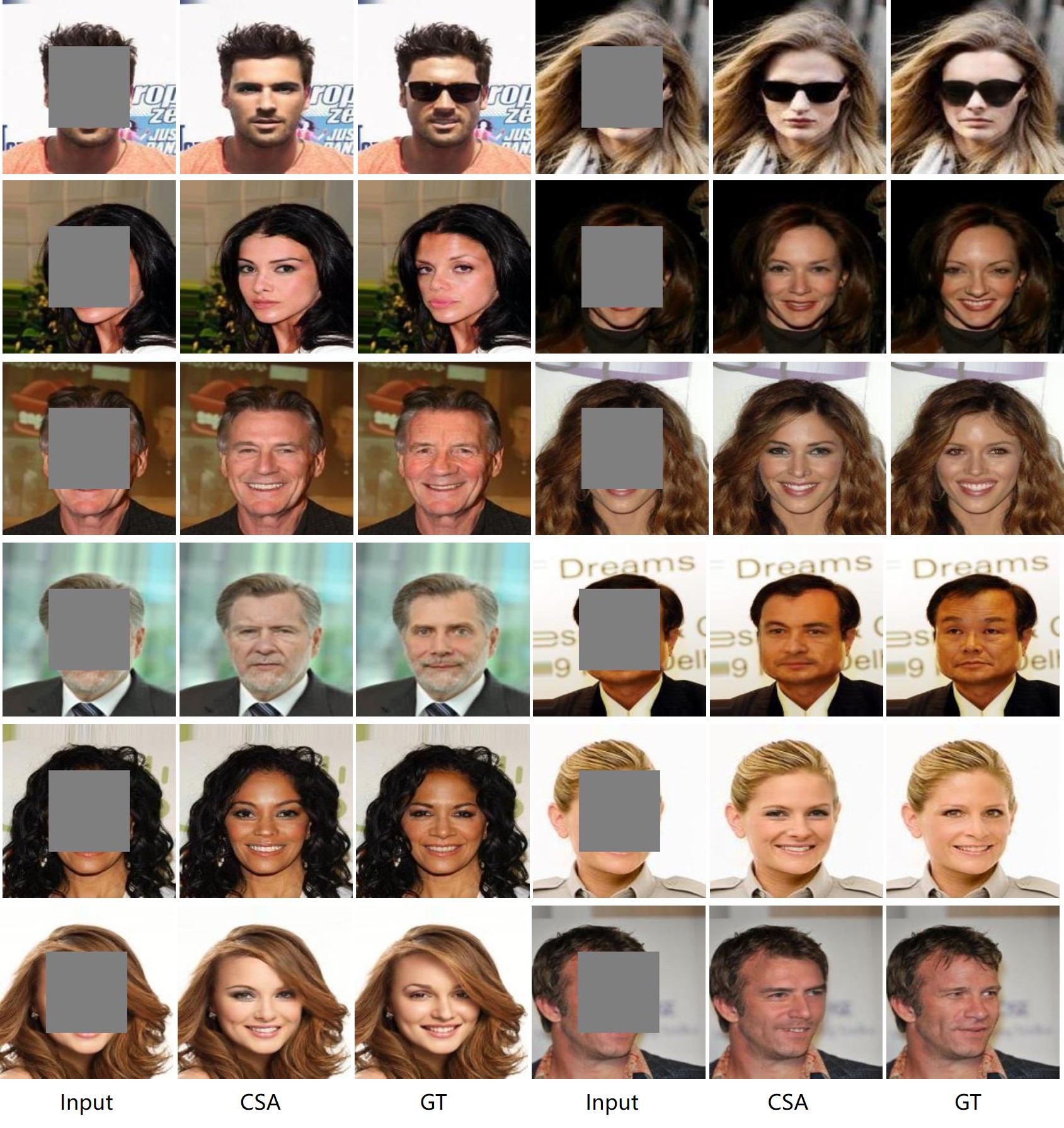}
\caption{More results on CelebA with centering masks.}
\label{imga5}
\end{figure*}

\begin{figure*}[t]
\centering
\includegraphics[scale=0.6]{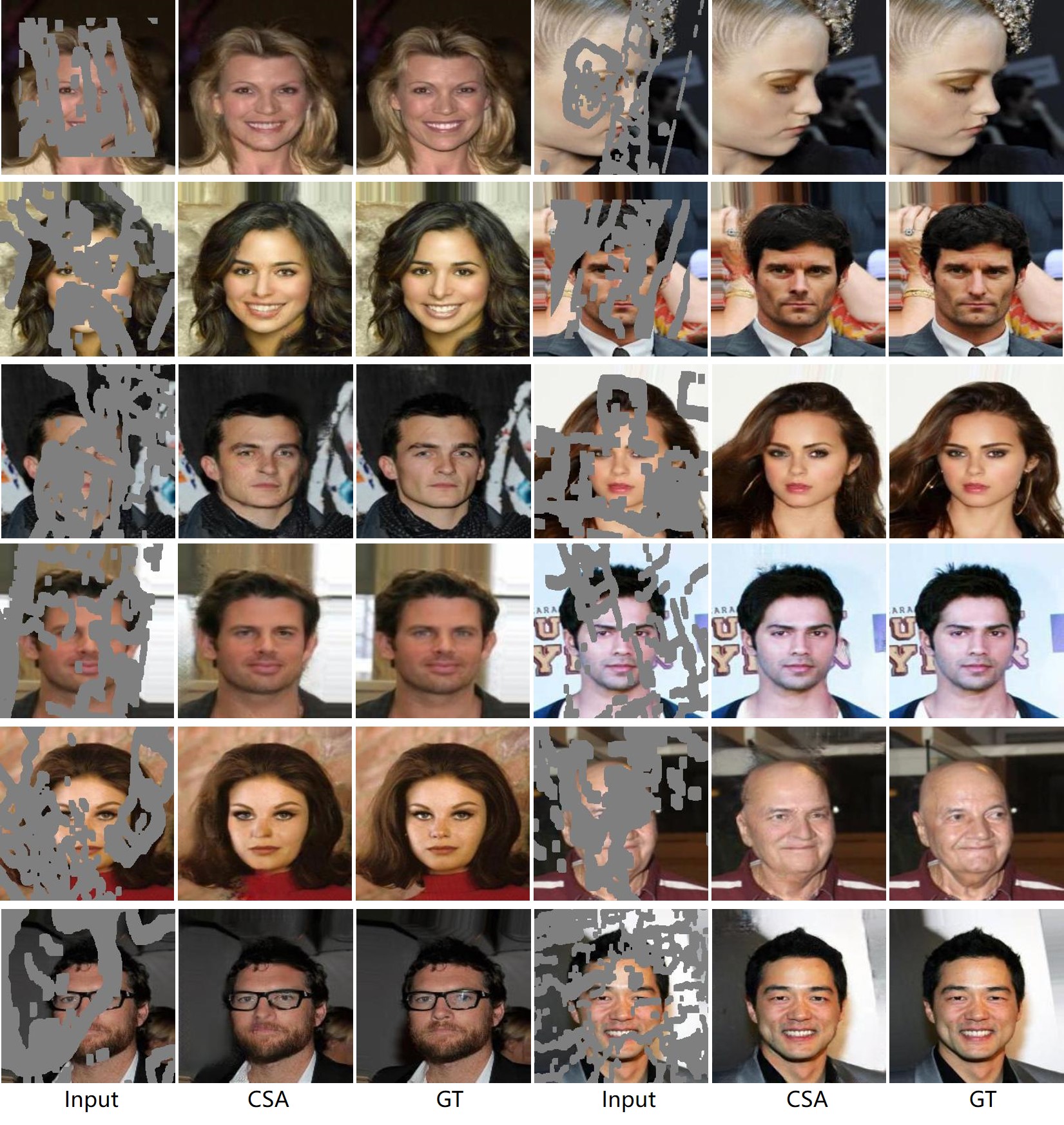}
\caption{More results on CelebA with irregular masks.}
\label{imga6}
\end{figure*}

\textbf{CelebA} Fig \ref{imga5} and Fig \ref{imga6} show more results obtained by our full model with centering and irregular masks respectively, where the model is trained on CelebA dataset. We resize image to 256$\times$256 for both training and evaluation.

\begin{figure*}[t]
\centering
\includegraphics[scale=0.6]{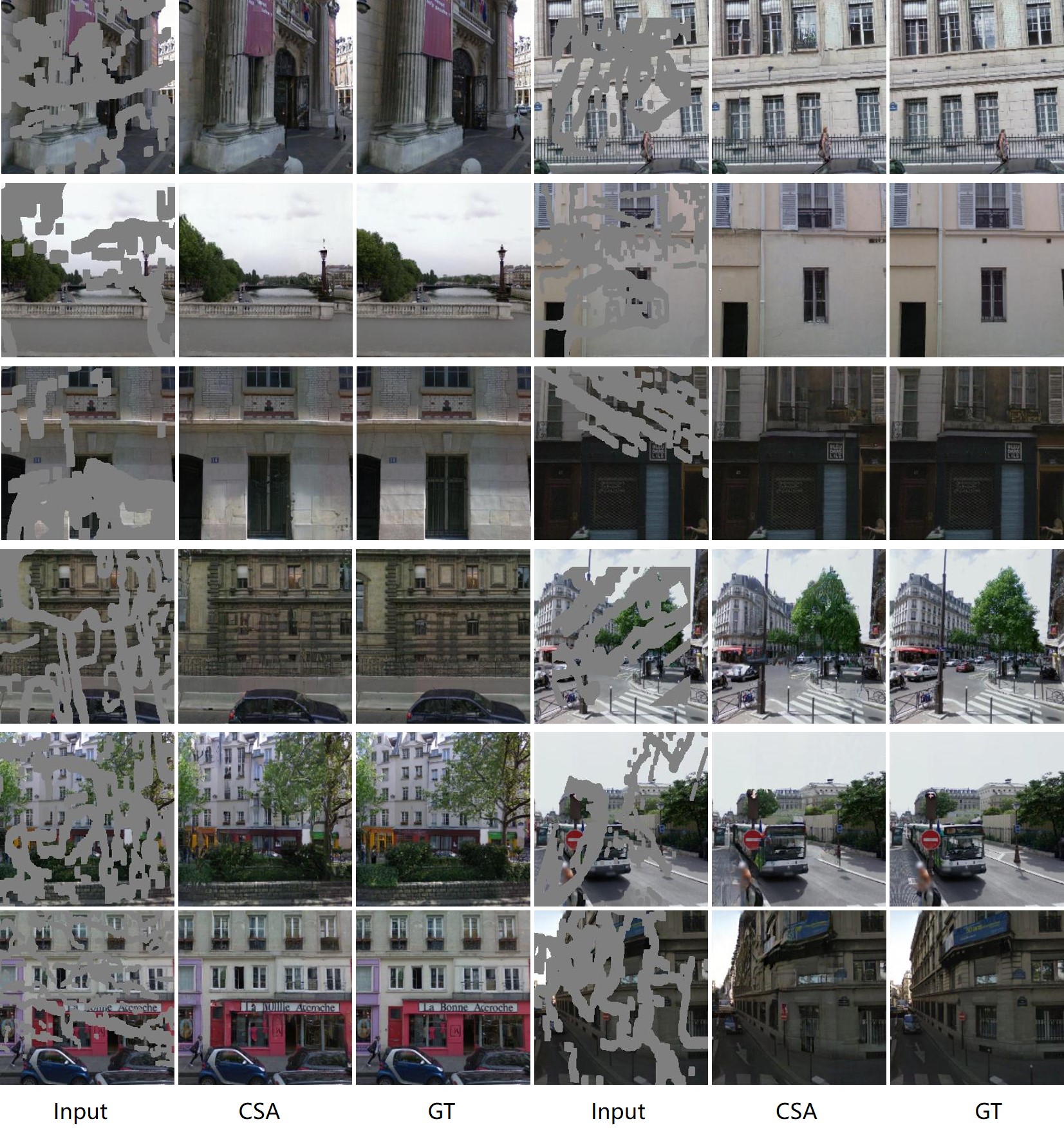}
\caption{More results on Paris StreetView with irregular masks.}
\label{imga7}
\end{figure*}

\textbf{Paris StreetView} We also perform experiments on our full model trained on Paris StreetView dataset with irregular masks, and the results are shown in Fig \ref{imga7}. We resize image to 256$\times$256 for both training and evaluation.

\begin{figure*}[t]
\centering
\includegraphics[scale=0.47]{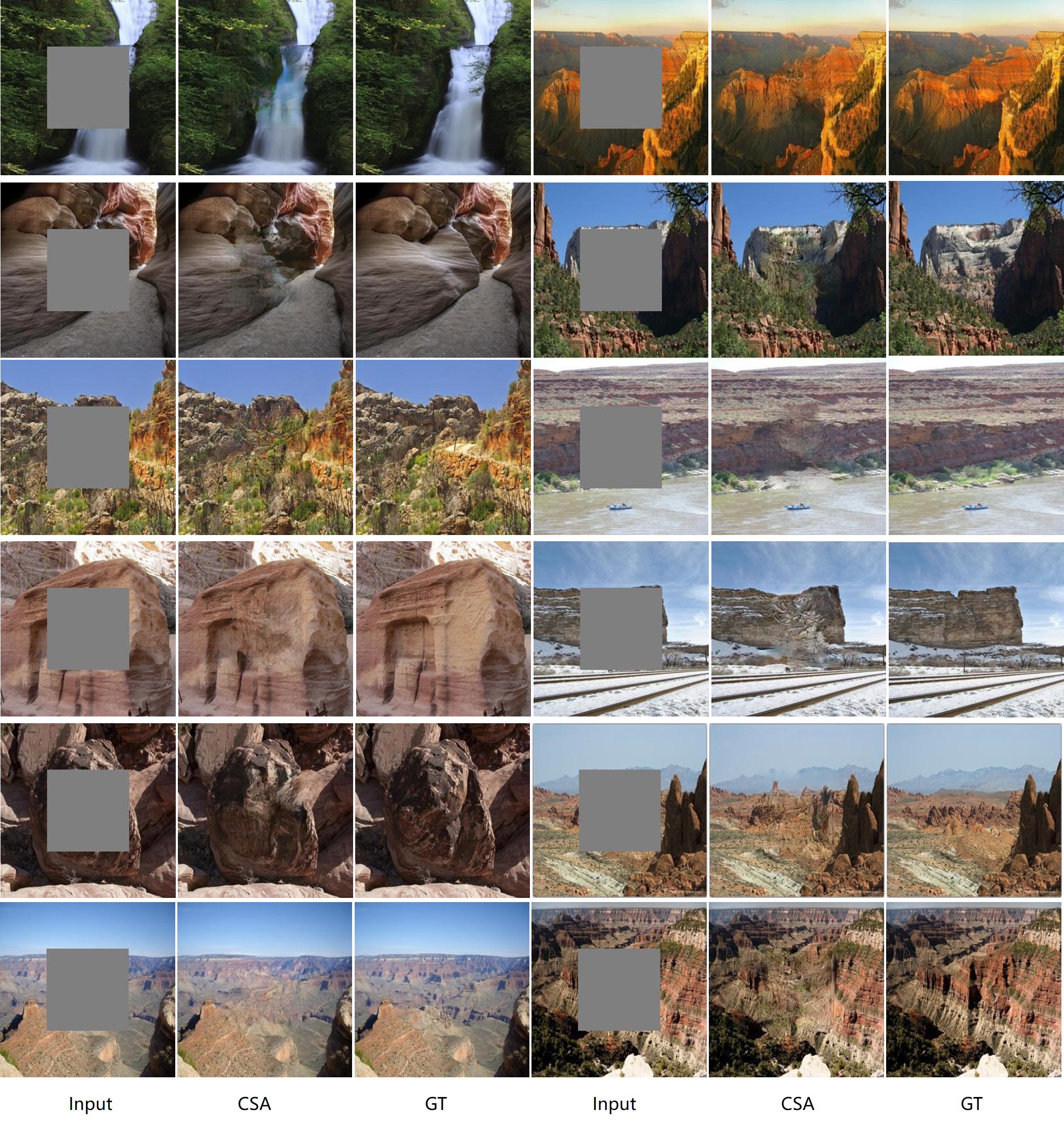}
\caption{More results on Place2 with centering masks.}
\label{imga8}
\end{figure*}

\textbf{Places2} Fig \ref{imga8} shows more results obtained by our full model with centering masks, where the model is trained on Places2 dataset. The scene categories selected from Places2 dataset are canyon and butte. We also resize the images to 256$\times$256 for both training and evaluation.

\begin{table*}[t]

\centering
 \begin{tabular}{|l|c|c|c|c|}
 \hline

 \textbf{The architecture of refinement network}          \\ \hline
  [Layer 1] Conv. (3, 3, 64), stride=1, padding=1;            \\\hline
  [Layer 2] LReLU; Conv. (4, 4, 64), stride=2, dilation=2, padding=3; IN;    \\
  LReLU; Conv. (3, 3, 128), stride=1, padding=1; IN; \\\hline
  [Layer 3] LReLU; Conv. (4, 4, 128), stride=2, dilation=2, padding=3; IN;      \\
  LReLU; Conv. (3, 3, 256), stride=1, padding=1; IN; \\\hline
  [Layer 4] LReLU; Conv. (4, 4, 256), stride=2, dilation=2, padding=3; IN;      \\
  LReLU; Conv. (3, 3, 512), stride=1, padding=1; \textbf{CSA}; IN; \\\hline
  [Layer 5] LReLU; Conv. (4, 4, 512), stride=2, dilation=2, padding=3; IN;  \\
    LReLU; Conv. (3, 3, 512), stride=1, padding=1; IN; \\\hline
  [Layer 6] LReLU; Conv. (4, 4, 512), stride=2, dilation=2, padding=3; IN;    \\
    LReLU; Conv. (3, 3, 512), stride=1, padding=1; IN; \\\hline
  [Layer 7] LReLU; Conv. (4, 4, 512), stride=2, dilation=2, padding=3; IN;     \\
    LReLU; Conv. (3, 3, 512), stride=1, padding=1; IN; \\\hline
  [Layer 8] LReLU; Conv. (4, 4, 512), stride=2, dilation=2, padding=3; IN;      \\
    LReLU; Conv. (3, 3, 512), stride=1, padding=1; IN; \\\hline
  [Layer 9] LReLU; Conv. (4, 4, 512), stride=2, padding=1;         \\\hline
  [Layer 10] ReLU; DeConv. (4, 4, 512), stride=2, padding=1; IN;    \\
  Concatenate(Layer 10, Layer 8); \\\hline
  [Layer 11] ReLU; DeConv. (3, 3, 512), stride=1, padding=1; IN; ; \\
  ReLU; DeConv. (4, 4, 512), stride=2, padding=1; IN;  \\
  Concatenate(Layer 11, Layer 7); \\\hline
  [Layer 12]  ReLU; DeConv. (3, 3, 512), stride=1, padding=1; IN;      \\
  ReLU; DeConv. (4, 4, 512), stride=2, padding=1; IN;  \\
  Concatenate(Layer 12, Layer 6); \\\hline
  [Layer 13]ReLU; DeConv. (3, 3, 512), stride=1, padding=1; IN;      \\
  ReLU; DeConv. (4, 4, 512), stride=2, padding=1; IN;  \\
  Concatenate(Layer 13, Layer 5); \\\hline
  [Layer 14] ReLU; DeConv. (3, 3, 512), stride=1, padding=1; IN;\\
  ReLU; DeConv. (4, 4, 512), stride=2, padding=1; IN;  \\
  Concatenate(Layer 14, Layer 4); \\\hline
  [Layer 15] ReLU; DeConv. (3, 3, 256), stride=1, padding=1; IN;     \\
  ReLU; DeConv. (4, 4, 256), stride=2, padding=1; IN;  \\
  Concatenate(Layer 15, Layer 3); \\\hline
  [Layer 16] ReLU; DeConv. (3, 3, 128), stride=1, padding=1; IN;       \\
  ReLU; DeConv. (4, 4, 128), stride=2, padding=1; IN;  \\
  Concatenate(Layer 16, Layer 2); \\\hline
  [Layer 17] ReLU; DeConv. (3, 3, 64), stride=1, padding=1; IN;    \\
  ReLU; DeConv. (4, 4, 64), stride=2, padding=1; IN;  \\
  Concatenate(Layer 17, Layer 1); \\\hline
  [Layer 18] ReLU; DeConv. (3, 3, 64), stride=1, padding=1;   \\\hline
 \end{tabular}

 \caption{The architecture of the refinement network. ¡°IN¡± represents InstanceNorm and ¡°LReLU¡± donates leaky ReLU with the slope of 0.2.
}
\label{taba4}
\end{table*}
\end{document}